\documentclass{article}
\usepackage{iclr2027_conference,times}

\usepackage{amsmath,amsfonts,bm}

\def\eqref#1{equation~\ref{#1}}
\def\1{\bm{1}}

\DeclareMathAlphabet{\mathsfit}{\encodingdefault}{\sfdefault}{m}{sl}
\SetMathAlphabet{\mathsfit}{bold}{\encodingdefault}{\sfdefault}{bx}{n}

\usepackage{hyperref}
\usepackage{url}
\usepackage{amsmath,amssymb,amsfonts}
\usepackage{graphicx}
\usepackage{booktabs}
\usepackage{xcolor}
\usepackage{enumitem}
\usepackage{algorithm}
\usepackage{algpseudocode}
\usepackage{microtype}
\usepackage{colortbl}
\usepackage{placeins}
\usepackage{float}
\usepackage{wrapfig}

\newcommand{\ourmethod}{\textsc{Stackelberg Alignment}}

\iclrfinalcopy
\begin{document}

\title{Multi-LLM Collaborative Alignment via Stackelberg Games}

\author{
Christina Hahn$^{1,*}$, Shangbin Feng$^{1,*}$, Dean Light$^{1}$, Swastik Roy$^{2}$, Hila Gonen$^{3}$, Yulia Tsvetkov$^{1}$ \\
$^{1}$University of Washington \quad $^{2}$Amazon \quad $^{3}$University of British Columbia \\
\texttt{chahn317@uw.edu}, \texttt{shangbin@cs.washington.edu}
}

\maketitle

\renewcommand{\thefootnote}{}\footnotetext{$^{*}$Equal contribution. Code: \url{https://github.com/BunsenFeng/model_collaboration}.}\addtocounter{footnote}{-1}\renewcommand{\thefootnote}{\arabic{footnote}}

\begin{abstract}
%Training-time model collaboration, where a pool of language models interact and produce collaborative training signals, is a promising paradigm for model alignment and evolution; yet existing methods treat instruction selection in the training curriculum as a fixed decision, sampling uniformly throughout training regardless of what the evolving model pool still needs to learn.
A pool of language models can collaborate and improve collectively by learning from one another's responses. These interactions depend on the instructions used during training. Existing methods typically sample instructions uniformly, even though their usefulness may change as the models improve: an instruction on which models' responses once differed in quality may later be answered equally well, while a previously difficult instruction may begin to provide a useful learning signal.
We propose \ourmethod{}, a game-theory-inspired leader--follower framework that turns instruction selection into an adaptive curriculum. %framing collaborative LLM alignment as a bilevel Stackelberg game: a bandit leader commits to an instruction distribution using EXP3, an adversarial bandit algorithm, before the follower models duel on those instructions, updating its leader weights with a reward combining instruction difficulty and duel discriminability.
%Each follower LLM maintains a reputation score updated from duel outcomes; models not in a duel serve as judges and their scores are weighted by reputation, with opponents matched by reputation proximity to keep duels competitive as the pool's skill levels diverge.
An EXP3 bandit acts as the leader, allocating a fixed sampling budget across instructions and updating its sampling distribution using a reward that combines instruction difficulty and response discriminability. The language models act as followers: they respond to the selected instructions, evaluate one another's responses, and learn from the resulting preference signals through DPO or GRPO. The framework uses Elo-style reputation-weighted peer judgment and reputation-based opponent matching to support reliable and competitive model interactions.
Experiments across three heterogeneous model pools and 12 benchmarks spanning scientific discovery, reasoning, code, instruction following, and knowledge show that \ourmethod{} achieves the highest macro-average across three diverse model pools, outperforming the strongest training-time baseline by up to 7.4\% and the best static inference baseline by 12--25\%.
Analysis confirms that the adaptive leader concentrates duels on the most informative instructions, and ablations show that both reputation-weighted judgment and reputation-based matching improve the effectiveness of multi-LLM evolution.
\end{abstract}

\section{Introduction}

Model collaboration allows language models to benefit from one another's complementary capabilities, and through inference-time interactions produce outputs better than any individual model~\citep{du2024position,feng2026one}.
Training-time collaboration offers a more fundamental opportunity: models can improve by interacting with an environment of other models, generating training signals that neither could produce alone.
Recent methods often achieve this through comparison and competition~\citep{luo2024arena,subramaniam2025multiagent}. Sparta Alignment~\citep{jiang2025sparta}, for example, has pairs of models respond to the same instruction, the remaining models judge the responses, and the winning pair is used for preference learning, reliably lifting all models in the pool across reasoning, code, and open-ended tasks. It further maintains reputation scores that weight peer judgments and guide opponent matching; we build on top of this framework. 
A key ingredient to all of these methods is, however, left unexplored: \emph{which instructions should models compete on}.
Existing approaches sample instructions uniformly throughout training with fixed schedule, rather than treating instruction selection as an adaptive component of the training loop.

Uniform instruction sampling is inefficient because instruction informativeness is non-stationary and heterogeneous.
Instructions where all models already agree yield no discriminative preference signal; instructions that all models fail on produce only tied scores with no learning gradient; the most valuable instructions are those at the \emph{frontier of disagreement}, where duels and pairwise comparisons generate learnable preference pairs.
This frontier shifts as models improve across iterations: an instruction that drives rich learning early in training may become trivial later, while harder instructions become tractable only after the pool has strengthened.
A fixed sampling distribution cannot track these shifts, wasting a large fraction of compute on uninformative instructions.
Adaptive instruction selection, informed by the evolving capabilities of the model pool, is therefore central to efficient collaborative alignment and training-time collaboration in general.

We frame this observation as a bilevel Stackelberg game~\citep{von1952theory,bacsar1998dynamic} between two classes of players.
A \textbf{leader} controls a distribution over the instruction pool and commits to it before any model generates a response; its objective is to concentrate compute on instructions that currently lie at the frontier of model disagreement.
The \textbf{followers} are the LLMs in the pool: they observe the selected instruction, duel, and are trained on the resulting preference pairs.
The leader implements this via EXP3~\citep{auer2002nonstochastic}, a no-regret bandit algorithm well suited to non-stationary reward sequences, with a composite reward that captures both instruction difficulty and duel discriminability.
Each model also maintains a reputation score reflecting its track record in past duels; models not part of a duel act as peer judges whose votes are weighted by their own reputations, and opponents are matched by reputation proximity so that duels remain competitive as skill levels diverge across training.
The resulting method, \ourmethod{}, is the first instantiation of collaborative LLM alignment with an adaptive leader that learns the instruction distribution jointly with the follower models, yielding a dynamic curriculum that concentrates compute on instructions where the pool still disagrees and shifts automatically as models strengthen.

\begin{figure*}[t]
  \centering
  \vspace*{-30pt}
  \includegraphics[width=0.9\linewidth]{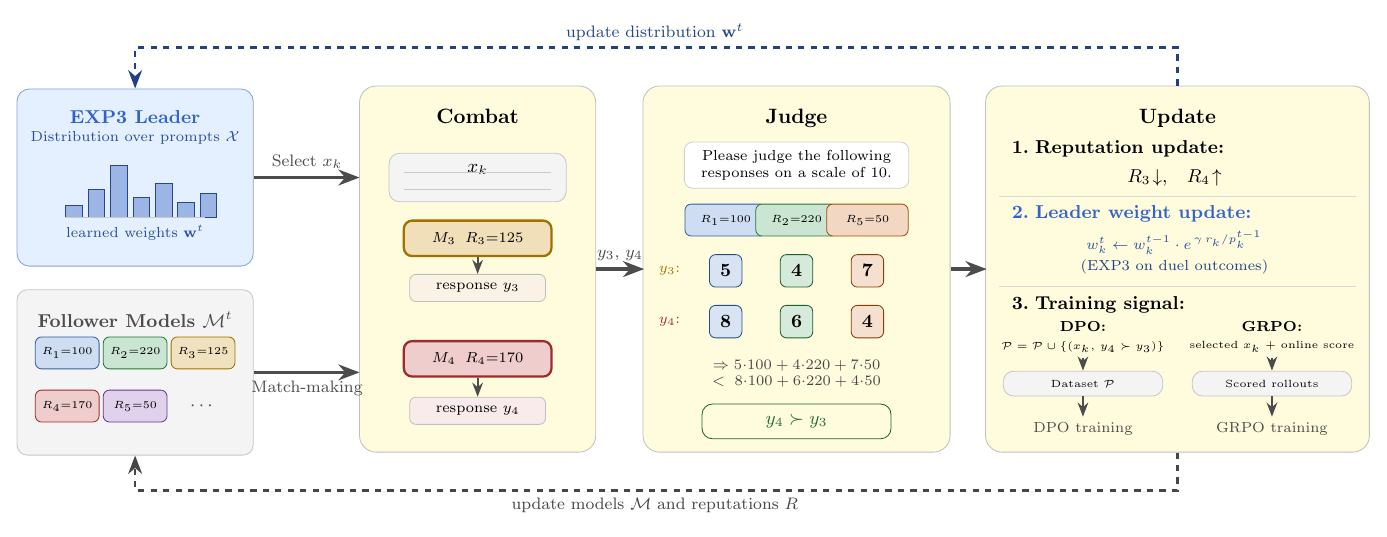}
  \vspace*{-10pt}
  \caption{Overview of \ourmethod{} for one iteration. The EXP3 leader maintains learned weights $\mathbf{w}^t$ over the instruction pool and selects instruction $x_k$ non-uniformly. Two follower models ($M_3$, $M_4$ in the figure) are matched by reputation and duel on $x_k$, generating responses $y_3$ and $y_4$. Remaining models act as peer judges, scoring both responses; scores are aggregated weighted by judge reputation, determining the winner. The Update step performs three actions: (1) reputation scores are adjusted, (2) leader weights are updated via EXP3 based on duel informativeness, and (3) the winning preference pair is added to the dataset $\mathcal{P}$ for DPO training, or the selected instruction and online judge scores are used directly as the reward for GRPO training. Dashed arrows show the resulting feedback loop: the updated leader weights reshape the instruction distribution, and the updated reputations and retrained models feed back into the follower pool for the next iteration.}
  \vspace*{-10pt}
  \label{fig:overview}
\end{figure*}

We evaluate \ourmethod{} on three pools of heterogeneous LLMs spanning specialized expert models, diverse academic research models, and general-purpose models, across 12 benchmarks covering scientific discovery, reasoning, code, instruction following, and knowledge.
\ourmethod{} achieves the highest macro-average in all three pools, outperforming other training-time algorithms by up to 7.4\% and the best static inference baseline by 12--25\%.
Our main contributions are: (1) a Stackelberg game formulation of training-time multi-LLM collaborative alignment; (2) an EXP3 bandit leader with a composite duel-informativeness reward that tracks the shifting capability frontier of the model pool; and (3) an empirical demonstration of consistent gains across three diverse model pools and twelve benchmarks, with analysis tracing the gains to the adaptive curriculum.

\section{Method}
\label{method}

Figure~\ref{fig:overview} illustrates one complete training iteration of \ourmethod{}.

\paragraph{Setup.}
Let $\mathcal{M}^0 = \{M_1^0, \ldots, M_m^0\}$ be a pool of $m$ language models and $\mathcal{X} = \{x_1, \ldots, x_K\}$ a dataset of instructions. Our goal is to produce an improved pool $\mathcal{M}^T$ by having models interact, evaluate, and learn from each other across $T$ iterations. Models engage in pairwise \emph{duels} on instructions: two models generate responses, the remaining models act as peer judges and score each response on a $[s_{\min}, s_{\max}]$ scale, and these scores are aggregated into preference pairs for RL fine-tuning. Each iteration runs $D$ such duels, with the instruction for each duel drawn independently (with replacement) from the leader's distribution $\pi_L$ rather than sweeping the instruction pool exhaustively; a given instruction may therefore be drawn multiple times, once, or not at all within a single iteration, and only duels that are actually drawn contribute preference pairs or reward signal to that iteration's training. This is an intentional design choice rather than an incidental side effect: concentrating duels on the currently most informative instructions, instead of exhausting the pool every iteration, is exactly the adaptive curriculum this framework is designed to realize (\S\ref{sec:leader}). Because the leader's sampling distribution keeps a nonzero floor probability for every instruction (Eq.~\ref{eq:exp3-dist}), no instruction is permanently excluded, though there is no guarantee that every instruction in the pool is drawn within any single iteration or even over the full training run. Following \citet{jiang2025sparta}, each model $M_i$ carries a \emph{reputation score} $R_i \in \mathbb{R}$ and a \emph{reputation deviation} $\sigma_i$, initialized uniformly. Reputation encodes each model's perceived reliability, as judged by its peers over training, and serves two roles: it governs opponent selection, and weights peer judgments during score aggregation.

\paragraph{A Stackelberg view of training-time model collaboration.}
Existing model collaboration approaches~\citep{du2024improving, subramaniam2025multiagent, wang2025mixture, jiang2025sparta} treat instruction selection as fixed, sampling uniformly from the training pool regardless of what the current model pool still needs to learn. However, instructions vary widely in difficulty and informativeness: some are already mastered by all models and yield no useful gradient, others are beyond reach and produce no meaningful preference signal, and the most valuable ones lie at the frontier of disagreement where duels generate genuine learning signal. Critically, as models collaborate and improve across iterations, this frontier shifts, making the optimal instruction distribution non-stationary. An adaptive curriculum that tracks these shifts is therefore essential for efficient collaborative alignment. We frame the duel-based training loop above as a \textbf{bilevel Stackelberg game}~\citep{von1952theory} between two classes of players.

The \textbf{leader} $L$ controls a distribution $\pi_L \in \Delta(\mathcal{X})$ over the instruction pool and commits to it before any model generates a response. Its strategic objective is to select the instructions that produce the most informative learning signal. The \textbf{followers} $\{M_i\}_{i=1}^m$ observe the selected instruction $x \sim \pi_L$ and respond by generating outputs, which are then evaluated and used for training. This sequential commitment structure (leader first, followers second) defines a Stackelberg game~\citep{bacsar1998dynamic}, and the leader's optimal strategy anticipates the followers' most informative responses.

This framework subsumes several prior alignment paradigms as special cases. When $\pi_L$ is fixed to uniform and $m = 1$, it recovers single-model self-alignment methods such as Self-Rewarding~\citep{yuan2025selfrewardinglanguagemodels} and SPIN~\citep{chen2024self}. When $\pi_L$ is uniform but $m > 1$ with pairwise competition, it recovers \textsc{Sparta Alignment}~\citep{jiang2025sparta}: we adopt its reputation-weighted peer judgment and Elo-style reputation updates directly (\S\ref{sec:reputation}), and extend its reputation-based opponent matching with a time-varying gap schedule (\S\ref{sec:matching}). \ourmethod{} is the first instantiation of the full framework with an \emph{adaptive} leader that learns $\pi_L$ jointly with the followers.

Algorithm~\ref{alg:main} summarizes the full procedure; the subsections below detail each component.

\begin{algorithm}[t]
\caption{\ourmethod{}}\label{alg:main}
\small
\begin{algorithmic}[1]
\Require Model pool $\mathcal{M}^0$, instruction set $\mathcal{X} = \{x_k\}_{k=1}^K$, iterations $T$, duels per iteration $D$
\State Initialize reputation $R_i \leftarrow R_0$, $\sigma_i \leftarrow \sigma_0$ for all $i$; leader weights $\mathbf{w}^0 \leftarrow \mathbf{1}_K$
\For{$t = 1, \ldots, T$}
  \State Judged duels $\mathcal{J} \leftarrow \emptyset$;\; preference pairs $\mathcal{P} \leftarrow \emptyset$
  \For{each duel $d = 1, \ldots, D$}
    \State $x \sim \pi_L(\mathbf{w}^{t-1})$ \Comment{Leader commits to instruction}
    \State Select active model $M_i^t$; draw opponent $M_{i'}^t \sim p_{\mathrm{match}}(\cdot \mid M_i^t, \{R_j\}, t)$ \Comment{Eq.~\ref{eq:opponent-selection}}
    \State \textbf{Combat:} generate $y_i \leftarrow M_i^t(x)$,\; $y_{i'} \leftarrow M_{i'}^t(x)$
    \State \textbf{Judge:} each non-combatant $M_k$ scores responses: $s_i^{(k)}, s_{i'}^{(k)} \in [s_{\min}, s_{\max}]$ \Comment{§\ref{sec:reputation}}
    \State $\bar{s}_i, \bar{s}_{i'} \leftarrow \mathrm{Aggregate}\bigl(\{s_i^{(k)}, s_{i'}^{(k)}\}_{k \notin \{i, i'\}}, \{R_k\}\bigr)$ \Comment{Eq.~\ref{eq:agg-score}}
    \State Add $(x, y_i, y_{i'}, \bar{s}_i, \bar{s}_{i'})$ to $\mathcal{J}$; update $R_i, R_{i'}, \sigma_i, \sigma_{i'}$ \Comment{Eq.~\ref{eq:rep-update}; see Appx.~\ref{app:implementation}}
    \State If $\bar{s}_i \neq \bar{s}_{i'}$: add preference pair to $\mathcal{P}$
  \EndFor
  \State \textbf{Leader update:} $\mathbf{w}^t \leftarrow \mathrm{EXP3\text{-}Update}(\mathbf{w}^{t-1}, \mathcal{J})$ \Comment{uses all judged duels}
  \State \textbf{Train (DPO):} $M_i^{t} \leftarrow \mathrm{DPO}(M_i^{t-1}, \mathcal{P})$ for all $i$
  \State \textbf{or Train (GRPO):} $M_i^{t} \leftarrow \mathrm{GRPO}(M_i^{t-1}, \{(x,y_i,y_{i'})\} \in \mathcal{J}, \{R_k\})$ for all $i$ \Comment{online peer rewards}
\EndFor
\State \Return Improved model pool $\mathcal{M}^T$
\end{algorithmic}
\end{algorithm}

\subsection{Stackelberg Leader: Adaptive Instruction Selection}
\label{sec:leader}

\paragraph{Instruction sampling distribution.}
At each training iteration $t \in \{1, \ldots, T\}$, the leader must pick which of the $K$ candidate instructions $\mathcal{X} = \{x_k\}_{k=1}^K$ to use for that iteration's duels; $p_k^t$ denotes the probability of sampling instruction $x_k$ at iteration $t$, and instructions that recently produced informative duels should be sampled more often. To this end, the leader maintains a weight vector $\mathbf{w}^t = (w_1^t, \ldots, w_K^t) \in \mathbb{R}_{>0}^K$ and induces the sampling distribution $p_k^t$ via EXP3~\citep{auer2002nonstochastic}:
\begin{equation}
    p_k^t = (1 - \gamma)\,\frac{w_k^t}{\|\mathbf{w}^t\|_1} + \frac{\gamma}{K},
    \label{eq:exp3-dist}
\end{equation}
where $\gamma \in [0,1]$ is the exploration rate. The $(1-\gamma)$ term exploits high-weight instructions while the $\gamma/K$ term encourages that all instructions are visited. EXP3 is fitting here because instruction rewards are non-stationary, meaning an instruction's informativeness is not fixed but drifts over the course of training: as models improve across iterations, the same instruction may yield rich signal early and none later. Many bandit algorithms assume each instruction's reward is drawn from a fixed distribution and can perform poorly once that assumption breaks; EXP3 instead offers a no-regret guarantee, i.e., its cumulative reward stays close to that of the best fixed instruction in hindsight, even against such adversarially shifting reward sequences.

\paragraph{Leader reward.}
After a duel on instruction $x_k$ with peer-aggregated scores $\bar{s}_i, \bar{s}_{i'} \in [s_{\min}, s_{\max}]$ (Eq.~\ref{eq:agg-score}), the leader receives a composite reward $r_k \in [0,1]$, which is fed into the EXP3 weight update below to raise or lower how often instruction $x_k$ is sampled in future iterations:

\emph{(i) Difficulty reward.} Among instructions that are learnable (score above a floor threshold), the leader prefers harder ones: instructions where models score perfectly are already mastered and offer no room for improvement, while those just above the threshold are most informative. For a model's aggregate score $s$ this is:
\begin{equation}
    r_{\mathrm{diff}}(s) = \begin{cases}
        \dfrac{s_{\max} - s}{s_{\max} - \tau_{\mathrm{th}}} & s \geq \tau_{\mathrm{th}} \\[4pt]
        0 & s < \tau_{\mathrm{th}}
    \end{cases}
    \label{eq:diff-reward}
\end{equation}
where $\tau_{\mathrm{th}}$ is a minimum score threshold below which the instruction is too hard to yield useful signal.

\emph{(ii) Preference quality reward.} For preference learning, the gap between chosen and rejected responses matters: deviations from the scheduled target in either direction are penalized: a gap of zero provides no learning signal, and a gap far above the target indicates the comparison is already trivially decided~\citep{yao2024varyingshadeswrongaligning}. We schedule a target gap $g^*_t$ that starts large (clear preference distinctions early) and decays (finer distinctions later):
\begin{equation}
    g^*_t = g_{\mathrm{end}} + (g_{\mathrm{start}} - g_{\mathrm{end}})\,(1 - \tau), \qquad \tau = \frac{t}{T-1},
    \label{eq:gap-schedule}
\end{equation}
and reward proximity to this target via a Gaussian kernel:
\begin{equation}
    r_{\mathrm{pref}}(\bar{s}_i, \bar{s}_{i'}) = \exp\!\left(-\frac{\bigl(|\bar{s}_i - \bar{s}_{i'}|\,/\,(s_{\max} - s_{\min}) - g^*_t\bigr)^2}{2\sigma_r^2}\right).
    \label{eq:pref-reward}
\end{equation}
where $\sigma_r$ is the kernel bandwidth. The instruction reward is computed per combatant and averaged: $r_k = \frac{1}{2}\sum_{s \in \{\bar{s}_i, \bar{s}_{i'}\}} \bigl(\lambda_1 r_{\mathrm{diff}}(s) + \lambda_2 r_{\mathrm{pref}}(\bar{s}_i, \bar{s}_{i'})\bigr)$, with $\lambda_1 + \lambda_2 = 1$.

\paragraph{EXP3 weight update.}
To correct for the bias introduced by non-uniform sampling, we apply the standard importance-weighted multiplicative update:
\begin{equation}
    \log w_k^{t+1} = \log w_k^t + \frac{1}{K} \cdot \frac{r_k \cdot \mathbf{1}[x_k \sim \pi_L^t]}{p_k^t},
    \label{eq:exp3-update}
\end{equation}
followed by re-normalization, where $p_k^t$ is from Eq.~\ref{eq:exp3-dist} and $\mathbf{1}[x_k \sim \pi_L^t]$ indicates whether instruction $x_k$ was sampled in the current duel. Instructions that yield high reward relative to their sampling probability receive increased weight, concentrating future sampling on informative prompts.

\subsection{Scheduled Follower Matching}
\label{sec:matching}

For each duel $d$, the active model $M_i^t$ is selected in round-robin order. Sparta Alignment~\citep{jiang2025sparta} draws its opponent from $\mathcal{M}^t \setminus \{M_i^t\}$ by reputation proximity; we extend this with a \emph{scheduled soft-matching} policy that targets a time-varying reputation gap, annealing from mismatched to near-peer opponents over training.

Define the normalized reputation gap between $M_i^t$ and any candidate $M_j$ as
\begin{equation}
    \hat{g}_{ij}^t = \frac{|R_i^t - R_j^t|}{\max_{k \neq i}\, |R_i^t - R_k^t|} \in [0, 1].
\end{equation}
We select the opponent by sampling from a distribution peaked at a target gap $g^*_t$:
\begin{equation}
    p(M_{i'}^t = M_j) \propto \exp\!\left(-\frac{(\hat{g}_{ij}^t - g^*_t)^2}{2\sigma_g^2}\right), \quad j \neq i,
    \label{eq:opponent-selection}
\end{equation}
where $\sigma_g$ is the selection bandwidth and $g^*_t$ follows Eq.~\ref{eq:gap-schedule} with $g_{\mathrm{start}}=1$, $g_{\mathrm{end}}=0$, giving $g^*_t = 1 - \tau$, which linearly decreases from 1 (early: mismatched opponents) to 0 (late: near-peer opponents).

This schedule has a curriculum interpretation. As reputations diverge over training, the early target of $g^* \approx 1$ steers the system toward mismatched opponents: the stronger model nearly always wins, producing clear preference pairs that provide clean supervision for improvement~\citep{sun2024easytohard}. Later, as the target shifts toward $g^* \approx 0$, near-peer duels produce finer-grained preference distinctions that drive precise refinement, analogous to competitive matchmaking in human rating systems~\citep{ebtekar2021elo}.

\subsection{Peer Judgment and Reputation System}
\label{sec:reputation}

Following \citet{jiang2025sparta}, all non-competing models evaluate both combat responses. Each judge $M_k^t \in \mathcal{M}^t \setminus \{M_i^t, M_{i'}^t\}$ assigns scores $s_i^{(k)}, s_{i'}^{(k)} \in [s_{\min}, s_{\max}]$ (a 1--10 scale, i.e., $s_{\min}=1$, $s_{\max}=10$) via an LLM-as-a-judge prompt~\citep{zheng2023judging}. The peer-aggregate score is a reputation-weighted mean:
\begin{equation}
    \bar{s}_i = \Bigl(\sum\nolimits_{k \notin \{i, i'\}} R_k^t\, s_i^{(k)}\Bigr) \Big/ \Bigl(\sum\nolimits_{k \notin \{i, i'\}} R_k^t\Bigr).
    \label{eq:agg-score}
\end{equation}
Weighting by $R_k^t$ discounts the influence of low-reputation judges, so that models with stronger track records contribute more to each score aggregation~\citep{wang2025creamconsistencyregularizedselfrewarding}.

\paragraph{Reputation update.}
Reputations are updated after each duel via an adapted Elo rule~\citep{ebtekar2021elo}, as in Sparta Alignment~\citep{jiang2025sparta}, that incorporates three factors:
\begin{equation}
    R_i \leftarrow R_i + (K_t/\lambda)\cdot
    \underbrace{\vphantom{\Big|}(\bar{s}_i - \bar{s}_{i'})}_{\text{\normalsize score gap}} \cdot
    \underbrace{\vphantom{\Big|}\tanh(\sigma_i)}_{\text{\normalsize stability}} \cdot
    \underbrace{\vphantom{\Big|}\max\!\Bigl(|\Phi(z_i) - \Phi(z_{i'})|,\;\epsilon\Bigr)}_{\text{\normalsize gap factor}},
    \label{eq:rep-update}
\end{equation}
where $z_i = (R_i - R_{i'})\big/\sqrt{\sigma_i^2 + \sigma_{i'}^2}$, $\Phi$ is the standard normal CDF, $\epsilon > 0$ is a floor preventing stagnation in near-tied duels, and $\lambda$ is a scaling constant. Reputations are floored at a fixed minimum to prevent degenerate negative drift. The $K$ factor decays exponentially over update count to cool large swings as reputations stabilize; $\sigma_i$ is the standard deviation of recent reputation deltas over a sliding window, measuring volatility. Full formulas are in Appendix~\ref{app:implementation}.

\subsection{Follower Training}
\label{sec:training}

\ourmethod{} supports two training settings for collaborative improvement. For DPO, non-tied preference pairs from all duels are collected and used for offline fine-tuning. For GRPO, the duel assignment structure determines each model's training prompts, with peer judges scoring completions online during training.

\paragraph{DPO.}
Non-tied preference pairs $(x, y_w \succ y_l)$ collected across all duels are used to fine-tune each model via Direct Preference Optimization~\citep{rafailov2024direct}, with the previous iteration's checkpoint as the reference policy, following a periodic hard-copy update schedule~\citep{gorbatovski2025learn}.

\paragraph{GRPO.}
As an online alternative, the duel structure determines each model's training prompts. Each model generates $G$ completions per assigned prompt; peer judges score each completion online, and the normalized peer score serves as the per-completion reward in the GRPO objective~\citep{shao2024deepseekmath}. Reputation updates are derived from the logged rewards after training completes.

\paragraph{Connection to the Stackelberg equilibrium.}
At convergence, the leader's instruction distribution $\pi_L^*$ maximizes expected instructional informativeness under the followers' best-response policies $\{\pi_i^*\}$; the followers in turn are optimal for the instruction distribution $\pi_L^*$ generates. This joint fixed point corresponds to a Stackelberg equilibrium~\citep{bacsar1998dynamic}, where neither the leader can improve its choice of instructions given the followers' behavior, nor can the followers improve their policies given the leader's instruction distribution. In practice we approximate this equilibrium via alternating updates: EXP3 for the leader and DPO/GRPO for the followers.

\section{Experiment Settings}
\label{experiment}

\paragraph{Models.}
We employ three model pools of varying scale and architecture for evaluation.

\textbf{Pool 1} contains 9 heterogeneous models with five Qwen2.5-7B variants (the base instruction-tuned model plus four domain fine-tunes on OASST1, ShareGPT, WizardLM, and biomedical corpora, in \cite{jiang2025sparta}), alongside AgentFlow-Planner-7B \citep{li2026flow}, Llama-3.1-8B-Instruct~\citep{llama3}, Aya-Expanse-8B \citep{dang2024aya}, and Pangea-7B \citep{yue2025pangea}. This pool spans diverse model families, training corpora, and areas of expertise.

\textbf{Pool 2} contains 8 models trained in diverse academic research projects, where researchers contribute specialized models as modular components to a collaborative system. We use the top-8 models sourced in \cite{feng2026scaling} (see Appendix~\ref{app:pools} for model identities).

\textbf{Pool 3} contains 4 general models: Qwen3.5-9B~\citep{qwen3.5}, Gemma-4-12B-Instruct~\citep{team2026gemma}, Nemotron-3-Nano-4B~\citep{blakeman2025nvidia}, and Phi-4~\citep{abdin2024phi}.

\paragraph{Baselines.}
We compare \ourmethod{} in two training configurations, \textbf{Stackelberg (DPO)} and \textbf{Stackelberg (GRPO)}, against eight baselines spanning static inference and training-based collaboration.

\textit{Static inference}: \textbf{Majority Vote}, ensemble plurality voting (applicable to objective tasks); \textbf{MoA}~\citep{wang2025mixture}, iterative synthesis and aggregation of multiple models' responses; \textbf{Multiagent Debate}~\citep{du2024improving}, iterative response refinement at inference time; \textbf{Heterogeneous Swarms}~\citep{feng2025heterogeneous}, which optimizes directed acylic graphs of model interactions for collaboration.
\textit{Training-based}: \textbf{Sparta Alignment}~\citep{jiang2025sparta}, the direct predecessor with a uniform instruction leader and DPO-only training; \textbf{Multiagent Fine-tuning}~\citep{subramaniam2025multiagent}, joint fine-tuning via debate-based data augmentation; \textbf{Trained Router}~\citep{ongroutellm}, which learns to route queries to the best model via supervised fine-tuning on validation-set oracle labels; \textbf{AggLM}~\citep{zhao2025majority}, a GRPO-trained aggregator with verifiable rewards.

\paragraph{Hyperparameters.}
For Stackelberg (DPO) and Stackelberg (GRPO), we run $T = 8$ iterations with $D = 64$ duels per iteration. For the EXP3 leader we set exploration rate $\gamma = 0.2$. The difficulty reward uses score threshold $\tau_{\mathrm{th}} = 3.0$, and the preference quality reward uses gap schedule $g_{\mathrm{start}} = 0.6$, $g_{\mathrm{end}} = 0.15$, $\sigma_r = 0.15$; the combined reward weights are $\lambda_1 = 0.3$ (difficulty) and $\lambda_2 = 0.7$ (preference quality). For opponent matching we use $\sigma_g = 0.15$ and random override probability $p_{\mathrm{rand}} = 0.2$, with a candidate pool of top-$3$ by reputation proximity. The reputation system uses $K_0 = 10$, $K_{\min} = 5$, decay rate $\alpha = 0.9$, decay step size $n_s = 10$, scaling factor $\lambda = 20$, and Elo floor $\epsilon = 0.01$. For DPO training, we use learning rate $1\times10^{-6}$, 1 epoch per iteration, and effective batch size 16 with LoRA. For GRPO training, we use the same learning rate and epoch schedule with $G = 4$ rollouts per prompt; all experiments run on NVIDIA H100 GPUs.

\paragraph{Datasets.}
We evaluate across 12 datasets in five domains.
\textit{Scientific} (4): BixBench~\citep{mitchener2025bixbench}, LabBench~\citep{laurent2024lab}, SMDD~\citep{han2026smdd}, and AssayBench~\citep{de2026assaybench}, spanning biology, bioinformatics, and drug discovery.
\textit{Reasoning} (2): GPQA-Diamond~\citep{rein2024gpqa} (graduate-level science) and MATH~\citep{hendrycks2021measuring}.
\textit{Code} (2): HumanEval~\citep{chen2021codex} and MBPP~\citep{austin2021program}.
\textit{Instruction following} (2): AlpacaEval~\citep{dubois2024alpacafarm} (score from a reward model, Skywork-Reward-Llama-3.1-8B-v0.2~\citep{liu2024skywork}) and IFEval~\citep{zhou2023instruction} (verifiable instruction-following accuracy).
\textit{Knowledge and truthfulness} (2): TruthfulQA~\citep{lin2022truthfulqa} and CulturalBench-Hard~\citep{chiu2024culturalbench}.

\begin{table*}[t]
\centering
\vspace*{-35pt}
\caption{Performance across three model pools and 12 datasets grouped by domain. \textbf{Bold}: best per column within each pool; \underline{underline}: second best; --: not applicable. Shaded rows (\colorbox{gray!12}{\strut}) are \ourmethod{} variants. The Avg column is the macro-average over all available datasets, with AlpacaEval min-max normalized to $[0,1]$.}
\label{tab:main}
\resizebox{0.9\textwidth}{!}{%
\begin{tabular}{l|cccc|cc|cc|cc|cc|c}
\toprule
& \multicolumn{4}{c|}{Scientific} & \multicolumn{2}{c|}{Reasoning} & \multicolumn{2}{c|}{Code} & \multicolumn{2}{c|}{Instruction} & \multicolumn{2}{c|}{Knowledge} & \\
Method & \rotatebox{75}{BixBench} & \rotatebox{75}{LabBench} & \rotatebox{75}{SMDD} & \rotatebox{75}{AssayBench} & \rotatebox{75}{GPQA-Dia} & \rotatebox{75}{MATH} & \rotatebox{75}{HumanEval} & \rotatebox{75}{MBPP} & \rotatebox{75}{AlpacaEval} & \rotatebox{75}{IFEval} & \rotatebox{75}{TruthfulQA} & \rotatebox{75}{CulturalBench} & \rotatebox{75}{Avg} \\
\midrule
\multicolumn{14}{c}{\textit{Pool 1: Specialized Expert LLMs}} \\
\midrule
Sparta Alignment & 0.301 & 0.308 & 0.168 & 0.017 & 0.313 & 0.820 & 0.719 & 0.634 & 7.528 & 0.707 & \textbf{0.686} & 0.656 & 0.524 \\
Majority Vote & 0.184 & 0.272 & -- & -- & 0.263 & 0.778 & -- & -- & -- & -- & 0.616 & 0.670 & 0.464 \\
AggLM & 0.291 & \underline{0.317} & 0.007 & \underline{0.029} & 0.273 & 0.817 & 0.649 & 0.565 & $-$2.647 & 0.469 & 0.605 & 0.591 & 0.407 \\
Trained Router & 0.233 & 0.263 & 0.011 & 0.015 & 0.313 & 0.735 & 0.623 & 0.511 & 1.718 & 0.631 & 0.605 & 0.633 & 0.428 \\
Multiagent FT & 0.223 & 0.316 & 0.187 & 0.019 & 0.273 & 0.729 & 0.640 & 0.579 & 5.785 & 0.625 & 0.585 & 0.682 & 0.475 \\
Multiagent Debate & \underline{0.319} & 0.292 & 0.011 & 0.015 & \underline{0.353} & 0.789 & 0.711 & 0.608 & $-$6.553 & 0.567 & 0.571 & 0.732 & 0.414 \\
Het.~Swarms & 0.194 & 0.264 & 0.000 & \textbf{0.033} & \textbf{0.364} & 0.789 & 0.597 & 0.495 & $-$3.633 & \underline{0.740} & 0.622 & \underline{0.746} & 0.420 \\
MoA & 0.223 & 0.287 & 0.095 & 0.016 & \underline{0.353} & 0.833 & \textbf{0.816} & 0.589 & $-$3.848 & 0.707 & 0.582 & 0.729 & 0.451 \\
\rowcolor{gray!12} Stackelberg (DPO) & 0.233 & 0.311 & \underline{0.204} & 0.024 & \textbf{0.364} & \underline{0.857} & \underline{0.763} & \underline{0.643} & \textbf{8.081} & \textbf{0.749} & 0.682 & \textbf{0.747} & \underline{0.548} \\
\rowcolor{gray!12} Stackelberg (GRPO) & \textbf{0.349} & \textbf{0.322} & \textbf{0.237} & 0.023 & \underline{0.353} & \textbf{0.877} & \textbf{0.816} & \textbf{0.653} & \underline{7.819} & 0.730 & \underline{0.684} & 0.724 & \textbf{0.563} \\
\midrule
\multicolumn{14}{c}{\textit{Pool 2: LLMs from Diverse Academic Research}} \\
\midrule
Sparta Alignment & 0.272 & 0.330 & \underline{0.216} & 0.019 & 0.303 & 0.853 & \underline{0.789} & 0.628 & \textbf{6.273} & 0.672 & 0.632 & 0.680 & 0.533 \\
Majority Vote & 0.146 & 0.287 & -- & -- & 0.263 & 0.825 & -- & -- & -- & -- & \underline{0.650} & 0.657 & 0.471 \\
AggLM & 0.282 & 0.282 & 0.011 & 0.015 & 0.283 & 0.727 & 0.439 & 0.382 & $-$1.835 & 0.416 & 0.551 & 0.582 & 0.341 \\
Trained Router & 0.214 & 0.261 & 0.170 & 0.015 & 0.343 & 0.826 & 0.702 & 0.616 & 3.066 & 0.649 & 0.614 & 0.654 & 0.476 \\
Multiagent FT & 0.233 & \underline{0.339} & 0.156 & 0.015 & 0.303 & 0.837 & 0.754 & 0.604 & 3.564 & 0.623 & 0.624 & 0.683 & 0.490 \\
Multiagent Debate & 0.280 & 0.311 & 0.096 & 0.014 & 0.313 & 0.805 & 0.693 & 0.622 & $-$2.921 & 0.583 & 0.574 & 0.353 & 0.387 \\
Het.~Swarms & \textbf{0.311} & 0.303 & 0.182 & \textbf{0.028} & 0.374 & \underline{0.884} & 0.693 & 0.598 & 0.440 & \textbf{0.721} & 0.638 & \underline{0.689} & 0.482 \\
MoA & 0.252 & 0.310 & 0.164 & 0.016 & \underline{0.414} & 0.826 & 0.737 & 0.616 & $-$0.524 & \underline{0.681} & 0.530 & 0.686 & 0.458 \\
\rowcolor{gray!12} Stackelberg (DPO) & \textbf{0.311} & 0.334 & 0.196 & \underline{0.026} & 0.393 & \textbf{0.889} & \textbf{0.816} & \underline{0.632} & 4.986 & 0.676 & \textbf{0.656} & \textbf{0.696} & \underline{0.540} \\
\rowcolor{gray!12} Stackelberg (GRPO) & \underline{0.291} & \textbf{0.355} & \textbf{0.234} & \underline{0.026} & \textbf{0.423} & 0.884 & \textbf{0.816} & \textbf{0.634} & \underline{5.207} & 0.680 & 0.626 & 0.680 & \textbf{0.544} \\
\midrule
\multicolumn{14}{c}{\textit{Pool 3: General-Purpose LLMs}} \\
\midrule
Sparta Alignment & \textbf{0.349} & 0.218 & 0.294 & \textbf{0.034} & 0.384 & 0.854 & \textbf{0.868} & 0.661 & \underline{7.109} & 0.775 & 0.770 & 0.749 & \underline{0.579} \\
Majority Vote & 0.214 & 0.171 & -- & -- & -- & -- & -- & -- & -- & -- & -- & -- & 0.192 \\
AggLM & 0.330 & \textbf{0.346} & 0.120 & \underline{0.031} & 0.313 & 0.857 & 0.535 & 0.550 & $-$2.975 & 0.569 & \textbf{0.789} & 0.790 & 0.436 \\
Trained Router & 0.243 & 0.173 & 0.073 & 0.016 & 0.292 & 0.825 & 0.781 & 0.606 & 5.888 & 0.589 & 0.681 & 0.667 & 0.485 \\
Multiagent FT & 0.340 & 0.263 & 0.139 & 0.027 & 0.208 & 0.494 & 0.447 & 0.413 & 3.603 & 0.655 & 0.682 & 0.683 & 0.417 \\
Multiagent Debate & 0.307 & 0.152 & 0.348 & 0.027 & 0.125 & \underline{0.890} & 0.614 & 0.626 & $-$1.965 & 0.765 & 0.707 & 0.830 & 0.458 \\
Het.~Swarms & 0.312 & 0.181 & 0.265 & 0.030 & 0.208 & 0.853 & \underline{0.860} & \textbf{0.691} & $-$2.623 & 0.736 & 0.357 & 0.803 & 0.444 \\
MoA & 0.291 & 0.257 & 0.303 & 0.024 & 0.375 & \textbf{0.906} & 0.500 & 0.626 & $-$1.233 & 0.769 & 0.770 & \textbf{0.853} & 0.487 \\
\rowcolor{gray!12} Stackelberg (DPO) & 0.322 & 0.309 & \textbf{0.390} & 0.030 & \underline{0.394} & 0.882 & \underline{0.860} & 0.663 & 4.106 & \underline{0.778} & 0.770 & 0.810 & 0.576 \\
\rowcolor{gray!12} Stackelberg (GRPO) & \underline{0.340} & \underline{0.330} & \underline{0.353} & 0.028 & \textbf{0.424} & 0.869 & \textbf{0.868} & \underline{0.684} & \textbf{7.138} & \textbf{0.791} & \underline{0.780} & \underline{0.841} & \textbf{0.609} \\
\bottomrule
\end{tabular}%
}
\vspace*{-15pt}
\end{table*}

\section{Results}

Table~\ref{tab:main} reports performance of all methods across the three model pools and 12 datasets.

\paragraph{Stackelberg consistently outperforms all baselines on average.}
\ourmethod{} achieves the highest macro-average across all three pools (Table~\ref{tab:main}). Stackelberg (GRPO) ranks first with Avg scores of 0.563, 0.544, and 0.609 in Pools~1--3, while Stackelberg (DPO) ranks second in Pools~1 and 2 (0.548 and 0.540). Relative to the strongest training-based baseline, Sparta Alignment, Stackelberg (GRPO) improves macro-average by 7.4\%, 2.1\%, and 5.2\% in the three pools. The gap over the best static inference baseline (MoA or Heterogeneous Swarms) is substantially larger, exceeding 20\% in Pools~1 and 3 and 12\% in Pool~2, confirming that (1) training-based collaboration methods are generally stronger and (2) \ourmethod{} further improves upon the state of the art by adaptively concentrating training on informative instructions. Appendix~\ref{app:significance} reports per-task 95\% confidence intervals for Stackelberg (GRPO) and (DPO), confirming that many of these individual gains are statistically significant.

\paragraph{Consistent gains on open-ended tasks.}
The most consistent column-wise advantage appears on SMDD (Small Molecule Drug Discovery), an open-ended scientific task: Stackelberg (GRPO) and (DPO) rank first and second on SMDD across all three pools. On AlpacaEval, Stackelberg achieves the highest reward-model score in Pools~1 and 3 (8.081 and 7.138). IFEval follows similarly, with Stackelberg (DPO) first in Pool~1 and Stackelberg (GRPO) first in Pool~3. A common thread is that these tasks are open-ended and lack a single correct answer, making the quality of the peer-judgment signal central to training progress; the EXP3 curriculum concentrates duels on instructions where the model pool still disagrees, producing richer preference pairs exactly where they matter most.

\paragraph{Gains transfer broadly across reasoning, code, and knowledge domains.}
The improvement extends well beyond the tasks most directly linked to preference learning. On GPQA-Diamond, Stackelberg (GRPO) leads in Pools~2 and 3 (0.423 and 0.424) and Stackelberg (DPO) ties for first in Pool~1 alongside Heterogeneous Swarms. On MATH, Stackelberg (GRPO) achieves the best score in Pool~1 (0.877) and Stackelberg (DPO) leads in Pool~2 (0.889); MATH instructions span a wide difficulty range, and we hypothesize this is exactly the setting where an adaptive curriculum that concentrates duels at the frontier of the pool's current ability (\S\ref{sec:leader}) has the most room to help relative to fixed uniform sampling. On HumanEval, Stackelberg (GRPO) ties for first in all three pools. Exceptions occur on narrow scientific tasks: LabBench in Pool~3 is led by AggLM, and AssayBench in Pools~1--2 is led by Heterogeneous Swarms. This suggests that highly specialized benchmarks like scientific discovery may benefit from more high-quality data in the instruction pool.

\paragraph{GRPO and DPO show complementary strengths.}
Within our two training variants, Stackelberg (GRPO) outperforms Stackelberg (DPO) on Avg in all pools and on most scientific and reasoning columns, while Stackelberg (DPO) is more competitive on instruction-following and cultural knowledge tasks (IFEval and CulturalBench in Pool~1; TruthfulQA and CulturalBench in Pool~2). Notably, Stackelberg (DPO) falls slightly below Sparta in Pool~3 (0.576 vs.\ 0.579), the one exception to Stackelberg's consistent advantage, suggesting that the GRPO objective provides a signal that DPO alone does not reliably deliver in this setting.

\paragraph{Margins vary with pool heterogeneity.}
Among non-Stackelberg training baselines, Sparta Alignment is consistently the strongest, validating the duel-based preference learning framework; the additional gains from the adaptive leader over Sparta range from 2.1\% to 7.4\%. The narrowest margin appears in Pool~2, where the heterogeneous academic training backgrounds of the models make it harder to find a single instruction distribution that benefits all models simultaneously. Pool~1 (specialized expert LLMs) shows the largest GRPO margin over Sparta (7.4\%), while Pool~3 (four general-purpose LLMs) shows the largest gap over static inference (25.1\% above MoA).

\vspace{-5pt}
\section{Analysis}
\vspace{-5pt}
\label{sec:analysis}

\paragraph{Pool heterogeneity and competitive stratification.}
\begin{wrapfigure}{r}{0.48\linewidth}
  \vspace{-1.0\baselineskip}
  \centering
  \vspace*{-35pt}
  \includegraphics[width=\linewidth]{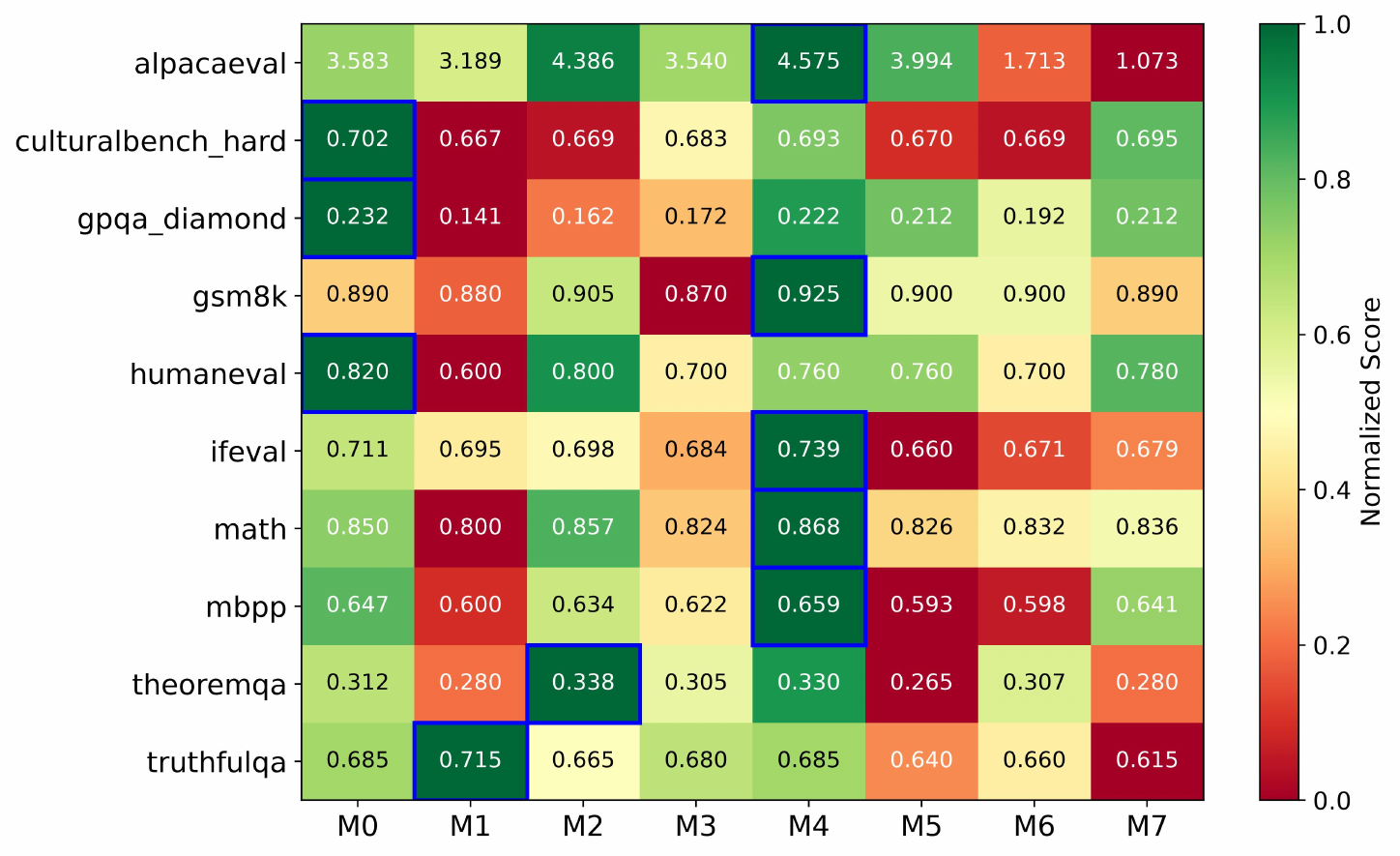}
  \vspace*{-20pt}
  \caption{Per-model eval scores at best iteration (Pool~2, Stackelberg GRPO). Row-normalized; green = best, red = worst per task.}
  \label{fig:specialization}
  \vspace*{-30pt}
\end{wrapfigure}

\begin{figure}[tb]
  \centering
  \vspace*{-30pt}
  \includegraphics[width=0.82\linewidth]{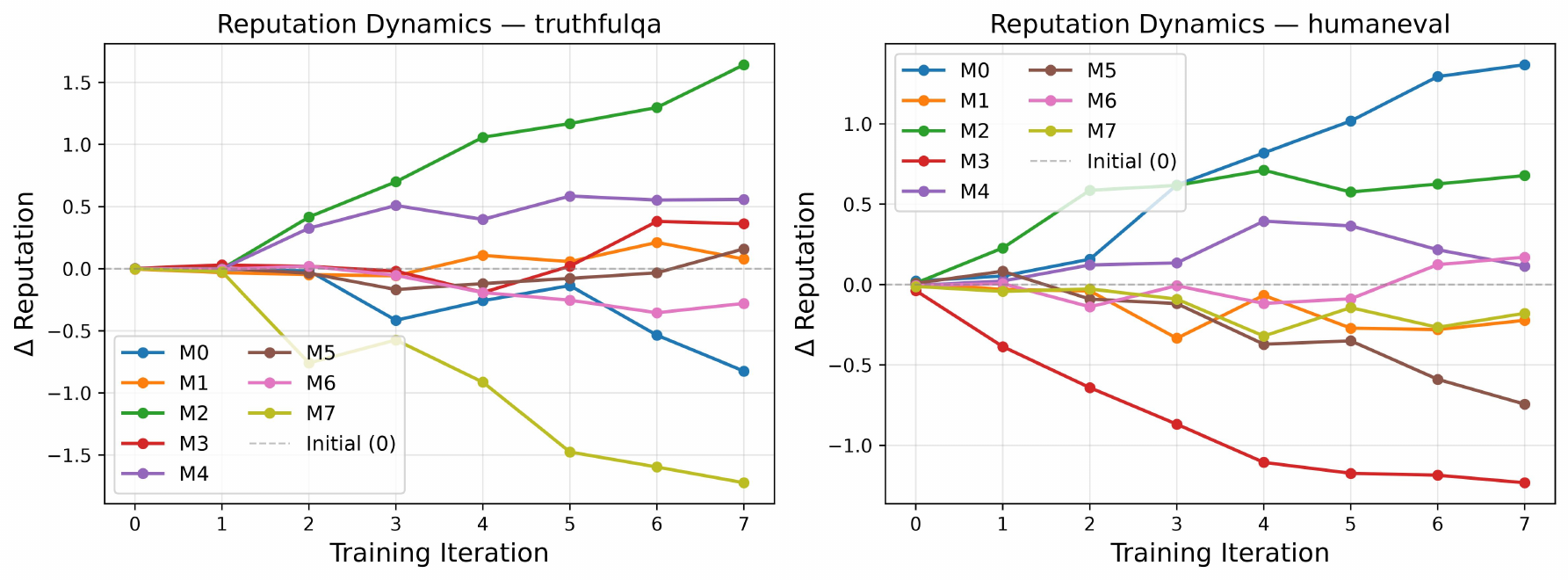}
  \vspace*{-10pt}
  \caption{Reputation score trajectories over training iterations for TruthfulQA and HumanEval (Pool~2, Stackelberg GRPO). Reputation scores diverge from the same initialization and stabilize by iteration~4--5. The leading model differs across tasks, confirming genuine task-specific competitive structure and collaborative learning landscape.}
  \label{fig:rating_dynamics}
  \vspace*{-15pt}
\end{figure}
\FloatBarrier
To achieve competitive pool training in \ourmethod{}, models must differ by capabilities. Figure~\ref{fig:specialization} (Pool~2, GRPO) confirms this: no model dominates all tasks, and the specialist structure is clear: \texttt{M4} leads on 5/10 tasks but loses to \texttt{M0} on reasoning and coding and to \texttt{M1} on TruthfulQA. This heterogeneity is captured by the reputation system: Figure~\ref{fig:rating_dynamics} shows reputation trajectories for TruthfulQA and HumanEval. Reputation scores gradually diverge from equal starting values, stabilize by iteration~4--5, and the winning model differs across tasks, providing evidence that task-specific competitive structure is resolved early and maintained throughout training.

\paragraph{Adaptive opponent selection.}

\begin{figure}[tb]
  \centering
  \vspace*{-40pt}
  \includegraphics[width=0.82\linewidth]{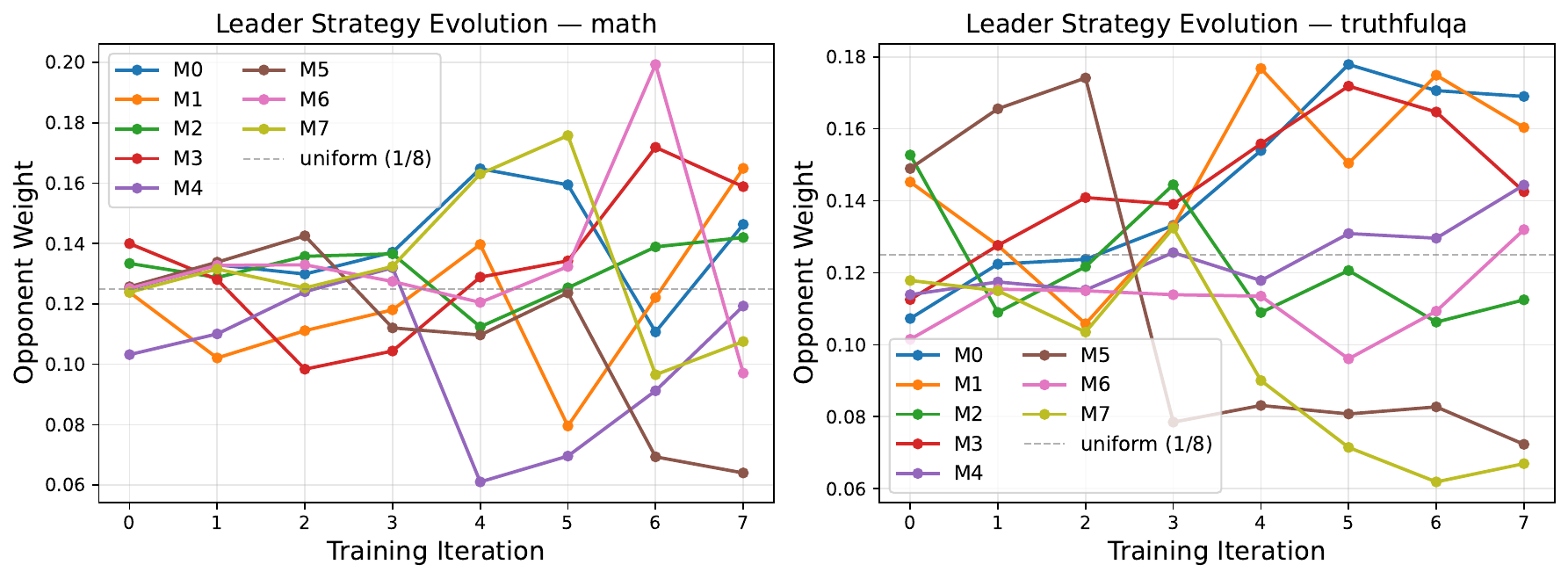}
  \vspace*{-10pt}
  \caption{Normalized opponent weight assigned by the Stackelberg leader over iterations (Pool~2, Stackelberg GRPO). Dashed line = uniform selection baseline. The leader shields the emerging dominant model from being excessively used as an opponent and adapts its strategy per task.}
  \label{fig:leader_strategy}
  \vspace{-1.0\baselineskip}
\end{figure}

\begin{figure}[tb]
  \centering
  \includegraphics[width=0.8\linewidth]{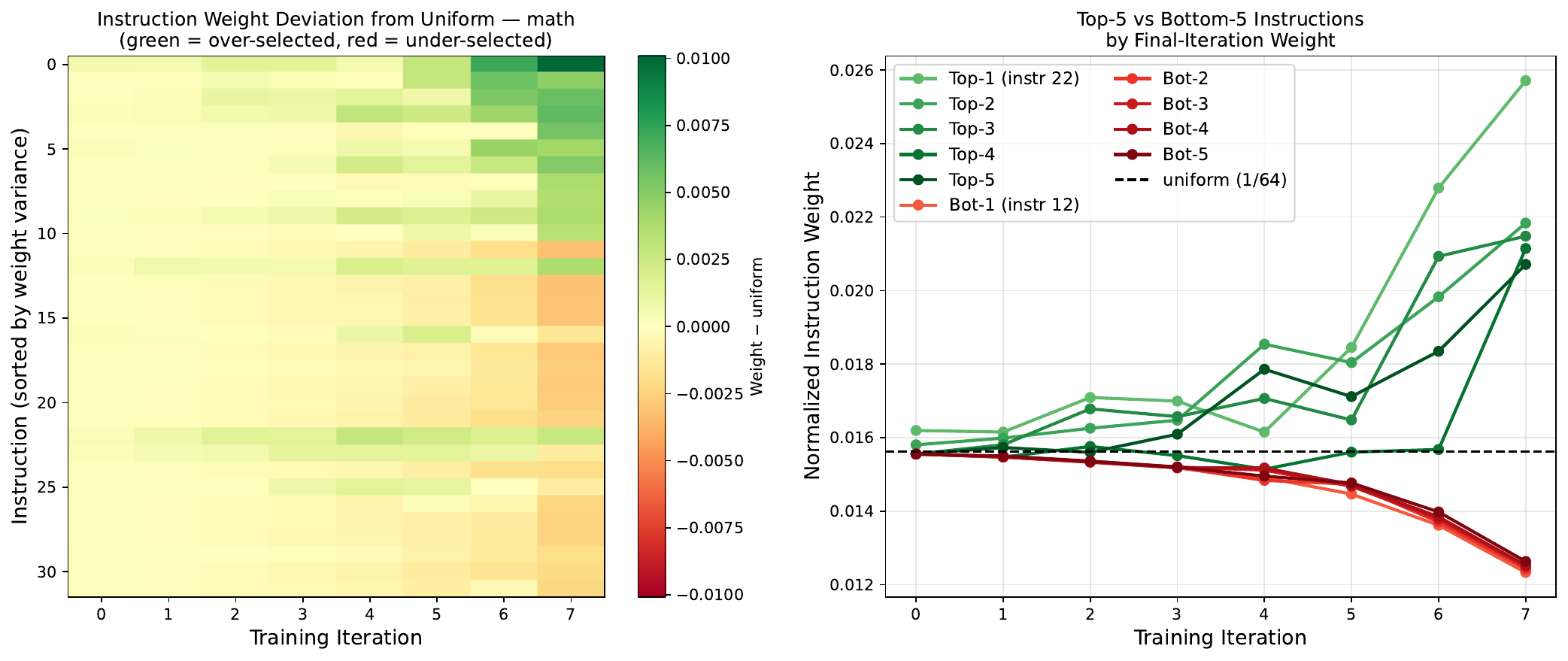}
  \vspace*{-10pt}
  \caption{Leader instruction weights over training iterations (the MATH dataset, model pool 1). Left: heatmap of normalized weight deviation from uniform for the top-32 most dynamic instructions. Right: trajectory of top-5 and bottom-5 instructions by final-iteration weight; dashed line = uniform. The leader converges to an implicit difficulty curriculum by iteration~5.}
  \label{fig:curriculum}
  \vspace{-1.0\baselineskip}
\end{figure}
\FloatBarrier

Given this heterogeneity, the competitive landscape the leader must navigate is non-trivial: pairwise win rates on MATH span 0.1--0.6 with no model dominating (Appendix~\ref{app:opponent_selection}, Figure~\ref{fig:winrate_matrix}), and the leader exploits this structure through adaptive opponent selection. Figure~\ref{fig:leader_strategy} traces the leader's normalized opponent weights over iterations for math and TruthfulQA. Early choices are already task-specific: math initially concentrates on \texttt{M3}; TruthfulQA peaks on \texttt{M5} at iteration~2. Notably, the model with the highest final eval score is actively shielded as an opponent at peak training time, with its weight dropping to $<\!0.5\times$ uniform at iterations 4--5, consistent with the Stackelberg objective of not wasting training budget on unwinnable duels. The strategy is genuinely task-adaptive: the same model (\texttt{M7}) is the most-selected opponent on math (weight $0.176$ at iteration~5) but nearly irrelevant on TruthfulQA (weight $0.062$ at iteration~7).

\paragraph{Emergent difficulty curriculum over instructions.}
Figure~\ref{fig:curriculum} shows the leader's per-instruction weights over training iterations (math task, Pool~1). Iterations 0--4 are near-uniform: the leader explores all instructions before committing. By iteration~5, the distribution sharpens: the top-5 instructions rise to $\sim$1.6$\times$ uniform by iteration~7, while the bottom-5 fall to $\sim$0.75$\times$ uniform, and the divergence exists and is accelerating. This constitutes an emergent difficulty curriculum: the leader identifies instructions where model disagreement persists, precisely where preference signal is most informative, and concentrates duels there without any explicit difficulty label. The behavior mirrors the leader's opponent-selection strategy (Fig.~\ref{fig:leader_strategy}): both start near-uniform and progressively concentrate attention as the competitive hierarchy becomes clearer.

\vspace{-5pt}
\section{Conclusion}
\vspace{-5pt}

We presented \ourmethod{}, a collaborative alignment framework for adaptive instruction selection and curriculum learning in multi-LLM evolution.
By casting the training loop as a Stackelberg game solved via EXP3, \ourmethod{} concentrates pairwise duels on the instructions where the model pool still disagrees.
Across three heterogeneous model pools and 12 benchmarks, \ourmethod{} with GRPO learning achieves the highest macro-average in every pool, improving over training-based baselines by up to 7.4\% and over the best static inference baseline by 12--25\%, with ablations confirming that the adaptive curriculum, reputation-weighted judgment, and reputation-based matching each contribute to these gains.

\subsection*{AI Use Statement}

We used generative AI tools to help with language polishing. The AI tools were used solely for language editing and did not contribute to the intellectual content of the work. We have reviewed all AI-assisted edits and take responsibility for the final content of this work.

\subsection*{Reproducibility Statement}

All hyperparameters are reported in Appendix~\ref{app:implementation} (Implementation Details), including EXP3 learning rate and reward coefficients, LoRA configuration, DPO and GRPO training settings, and generation parameters. Dataset sizes and splits are provided in Appendix~\ref{app:data} (Dataset Statistics). Baseline configurations are listed in Appendix~\ref{app:baselines} (Baseline Configurations), and the judge prompt is presented in Appendix~\ref{app:judge_prompt} (Judge Prompt). Statistical significance for our main results is reported in Appendix~\ref{app:significance}. Our method builds on publicly available pretrained models and standard training libraries; no proprietary data or infrastructure is required beyond the compute described in Appendix~\ref{app:implementation}.

\subsection*{Ethics Statement}

\ourmethod{} is designed to improve language model alignment through collaborative training: models improve each other via structured competition and peer judgment, which is broadly beneficial. However, competitive training pools also introduce potential misuse scenarios. A malicious actor who controls one model in the pool could manipulate duel outcomes to inflate its own reputation score, skew peer judgment, and inject adversarial preference pairs into the training data, degrading the alignment of all other models in the pool~\citep{yang2026among}. The reputation system provides partial robustness by down-weighting low-reputation judges, but does not make it a formal defense against adversarial participants. We see adversarial robustness of collaborative training as an important open problem, particularly as decentralized or federated model collaboration systems are deployed at scale.

\bibliography{iclr2027_conference}
\bibliographystyle{iclr2027_conference}

\appendix
\section{Related Work}

\paragraph{Model collaboration.}
The landscape of multi-LLM collaboration can be organized by the level of information exchange~\citep{feng2026scaling}.
API-level methods route or cascade queries to the most suitable model in the pool~\citep{ongroutellm,fenggraphrouter,chenfrugalgpt}.
Text-level methods have models exchange generated text: through debate and multi-round discussion~\citep{du2024improving}, response synthesis across model layers~\citep{wang2025mixture}, structured interaction graphs~\citep{feng2025heterogeneous}, and modular value alignment~\citep{feng2024modular}.
Weight-level methods operate directly in parameter space via model merging.
Among training-based approaches, Multiagent Fine-tuning~\citep{subramaniam2025multiagent} uses debate-generated data for supervised fine-tuning; AggLM~\citep{zhao2025majority} trains an RL aggregator over model outputs; Arena Learning~\citep{luo2024arena} simulates pairwise ELO-rated battles to generate post-training data; and Sparta Alignment~\citep{jiang2025sparta} trains models through competitive duels with peer judgment.
\ourmethod{} extends this line by introducing an adaptive leader that determines \emph{which instructions} drive collaborative training at each step, turning the instruction distribution itself into a learned, non-stationary curriculum.

\paragraph{Self-play and curriculum learning.}
A related line of work trains LLMs through self-play and iterative self-improvement.
SPIN~\citep{chen2024self} and Self-Rewarding LMs~\citep{yuan2025selfrewardinglanguagemodels} improve a single model by having it compete against or judge its own prior outputs.
SPIRAL~\citep{liu2026spiral} extends self-play to zero-sum multi-agent games, generating an automatic opponent curriculum; SPADE~\citep{liu2026spade} trains a single model as both environment designer and reasoning agent, adaptively targeting its own capability frontier.
A parallel thread develops curriculum learning for LLM fine-tuning: easy-to-hard generalization~\citep{sun2024easytohard} shows that ordering training examples by difficulty improves reasoning; SEC~\citep{chen2025self} frames curriculum selection as a non-stationary bandit over instruction categories; CurES~\citep{zeng2026cures} uses gradient signals to identify the most informative training examples at each stage.
\ourmethod{} unifies these threads: it applies bandit-driven instruction selection within a competitive multi-LLM training loop, where the difficulty signal emerges from inter-model disagreement rather than from predefined categories or gradient heuristics.

\paragraph{Game-theoretic alignment.}
Our work is grounded in game-theoretic alignment, where the interaction between alignment objectives and model responses is formalized as a strategic game between rational agents rather than a fixed optimization target.
Nash Learning from Human Feedback~\citep{pmlr-v235-munos24a} formalizes preference optimization as a two-player constant-sum game with the Nash policy as solution; INPO~\citep{zhang2025iterative} iteratively approximates the Nash policy via no-regret online learning.
These Nash-based methods model alignment as a symmetric simultaneous game.
\ourmethod{} instead adopts a Stackelberg formulation~\citep{von1952theory,bacsar1998dynamic}: the leader (instruction selector) commits to a distribution before the followers (LLMs) respond and train on such data, capturing the natural asymmetry between curriculum design and model adaptation.
SSAPO~\citep{xu2026stackelberg} also applies Stackelberg games to LLM alignment, but as a two-policy robustness problem against noisy preference labels; \ourmethod{} applies the framing to a multi-follower competitive pool where the leader maximizes instructional informativeness.
Bilevel RL for incentive alignment~\citep{thoma2024contextual} provides further theoretical grounding for leader-follower RL under best-response constraints.

\section{Additional Results}

\subsection{Statistical Significance}
\label{app:significance}

Table~\ref{tab:ci} reports per-task 95\% confidence intervals for Stackelberg (GRPO) and (DPO) across all three pools, computed via 10,000 Monte Carlo draws (Bernoulli for binary tasks; Gaussian for AlpacaEval with $\sigma = 5.0/\sqrt{n}$). A $^*$ mark indicates the point estimate exceeds every baseline on that task; $^{**}$ indicates the CI lower bound does. Stackelberg (GRPO) earns at least one $^{**}$ mark in every pool, and at least one $^*$ mark on 7--8 of the 10 reported tasks. SMDD and AssayBench are excluded: their small test sets ($n \leq 334$) inflate Gaussian CIs to $\pm 0.55$, making interval comparisons uninformative.

\begin{table*}[h]
\centering
\small
\caption{Per-task scores and 95\% confidence intervals for Stackelberg (GRPO) and Stackelberg (DPO). CIs computed via 10,000 Monte Carlo draws (Bernoulli for binary tasks; Gaussian for AlpacaEval). $^*$: point estimate exceeds every baseline. $^{**}$: CI lower bound exceeds every baseline. SMDD and AssayBench omitted ($\text{CI} \approx \pm 0.55$).}
\label{tab:ci}
\begin{tabular}{l rr rr}
\toprule
 & \multicolumn{2}{c}{Stackelberg (GRPO)} & \multicolumn{2}{c}{Stackelberg (DPO)} \\
\cmidrule(lr){2-3}\cmidrule(lr){4-5}
Task & Score & 95\% CI & Score & 95\% CI \\
\midrule
\multicolumn{5}{l}{\textit{Pool 1: Specialized Expert LLMs}} \\
\midrule
BixBench      & $0.350^{*}$    & [0.262, 0.447] & $0.233$         & [0.155, 0.320] \\
LabBench      & $0.322^{*}$    & [0.284, 0.360] & $0.311$         & [0.273, 0.348] \\
GPQA-Dia      & $0.354$        & [0.263, 0.444] & $0.364$         & [0.273, 0.455] \\
MATH          & $0.877^{**}$   & [0.856, 0.898] & $0.857^{**}$    & [0.834, 0.879] \\
HumanEval     & $0.816$        & [0.746, 0.886] & $0.763$         & [0.684, 0.842] \\
MBPP          & $0.653^{*}$    & [0.610, 0.696] & $0.643$         & [0.602, 0.684] \\
AlpacaEval    & $7.819^{*}$    & [7.512, 8.127] & $8.081^{**}$    & [7.771, 8.397] \\
IFEval        & $0.730$        & [0.676, 0.784] & $0.749^{*}$     & [0.692, 0.800] \\
TruthfulQA    & $0.684$        & [0.647, 0.720] & $0.682$         & [0.645, 0.718] \\
CulturalBench & $0.724$        & [0.687, 0.759] & $0.747^{*}$     & [0.711, 0.781] \\
\midrule
\multicolumn{5}{l}{\textit{Pool 2: Diverse Academic Research}} \\
\midrule
BixBench      & $0.291$        & [0.204, 0.379] & $0.311$         & [0.223, 0.398] \\
LabBench      & $0.355^{*}$    & [0.317, 0.395] & $0.334$         & [0.296, 0.374] \\
GPQA-Dia      & $0.423^{*}$    & [0.323, 0.515] & $0.393$         & [0.293, 0.495] \\
MATH          & $0.884$        & [0.863, 0.904] & $0.889^{*}$     & [0.868, 0.909] \\
HumanEval     & $0.816^{*}$    & [0.746, 0.886] & $0.816^{*}$     & [0.746, 0.886] \\
MBPP          & $0.634$        & [0.591, 0.678] & $0.632$         & [0.589, 0.674] \\
AlpacaEval    & $5.207$        & [4.897, 5.514] & $4.986$         & [4.676, 5.303] \\
IFEval        & $0.680$        & [0.620, 0.736] & $0.676$         & [0.620, 0.732] \\
TruthfulQA    & $0.626$        & [0.588, 0.663] & $0.656$         & [0.618, 0.692] \\
CulturalBench & $0.680$        & [0.643, 0.718] & $0.696^{*}$     & [0.659, 0.731] \\
\midrule
\multicolumn{5}{l}{\textit{Pool 3: General-Purpose LLMs}} \\
\midrule
BixBench      & $0.340$        & [0.252, 0.437] & $0.322$         & [0.233, 0.418] \\
LabBench      & $0.330$        & [0.291, 0.370] & $0.309$         & [0.272, 0.346] \\
GPQA-Dia      & $0.424$        & [0.323, 0.525] & $0.394$         & [0.303, 0.495] \\
MATH          & $0.869$        & [0.848, 0.890] & $0.882$         & [0.861, 0.902] \\
HumanEval     & $0.868$        & [0.807, 0.930] & $0.860$         & [0.790, 0.921] \\
MBPP          & $0.684$        & [0.641, 0.725] & $0.663$         & [0.620, 0.706] \\
AlpacaEval    & $7.138$        & [6.830, 7.449] & $4.106$         & [3.796, 4.417] \\
IFEval        & $0.791^{*}$    & [0.740, 0.840] & $0.778^{*}$     & [0.724, 0.828] \\
TruthfulQA    & $0.780$        & [0.746, 0.812] & $0.770$         & [0.736, 0.802] \\
CulturalBench & $0.841$        & [0.811, 0.870] & $0.810$         & [0.778, 0.842] \\
\bottomrule
\end{tabular}
\end{table*}

\subsection{Cross-Task Transfer}

Table~\ref{tab:transfer} tests whether competitive pool training on one task transfers to others. We take the best Stackelberg (GRPO) checkpoint trained exclusively on MATH-type instructions (Pool~2, \texttt{parti\_4\_full}, iteration~6) and evaluate it on HumanEval and MBPP; symmetrically, we take the best checkpoint trained on HumanEval instructions (\texttt{parti\_0\_full}, iteration~7) and evaluate it on MATH and MBPP. The baseline is the best untrained single model in the pool per task.

\begin{table}[h]
\centering
\small
\caption{Cross-task transfer in Pool~2. Each trained column shows the best single-model checkpoint from competitive pool training on that source task, evaluated on the target. Baseline: best untrained model in the pool per task. ``No gain'' indicates same-task training matched or fell below the untrained baseline.}
\label{tab:transfer}
\begin{tabular}{l rrrr}
\toprule
 & & \multicolumn{3}{c}{Checkpoint source} \\
\cmidrule(lr){3-5}
Eval Task & Best Untrained & Same-Task & Math-Trained & HumanEval-Trained \\
\midrule
MATH      & 0.863 & 0.865 & {--}                  & 0.866 \\
HumanEval & 0.790 & 0.790 (no gain) & \textbf{0.974} {(+18pp)} & {--} \\
MBPP      & 0.639 & 0.632 (no gain) & \textbf{0.963} {(+33pp)} & \textbf{0.961} {(+32pp)} \\
\bottomrule
\end{tabular}
\end{table}

Training on MATH transfers dramatically to coding (+18 pp on HumanEval, +33 pp on MBPP), while same-task pool training on coding yields no improvement over the best untrained model. HumanEval pool training likewise transfers to MBPP (+32 pp). These results suggest that competitive training on mathematical reasoning produces broadly transferable gains rather than task-specific memorization.

\subsection{Ablation: Opponent Selection Strategy}
\label{app:opponent_selection}

\begin{figure}[t]
  \centering
  \includegraphics[width=0.55\linewidth]{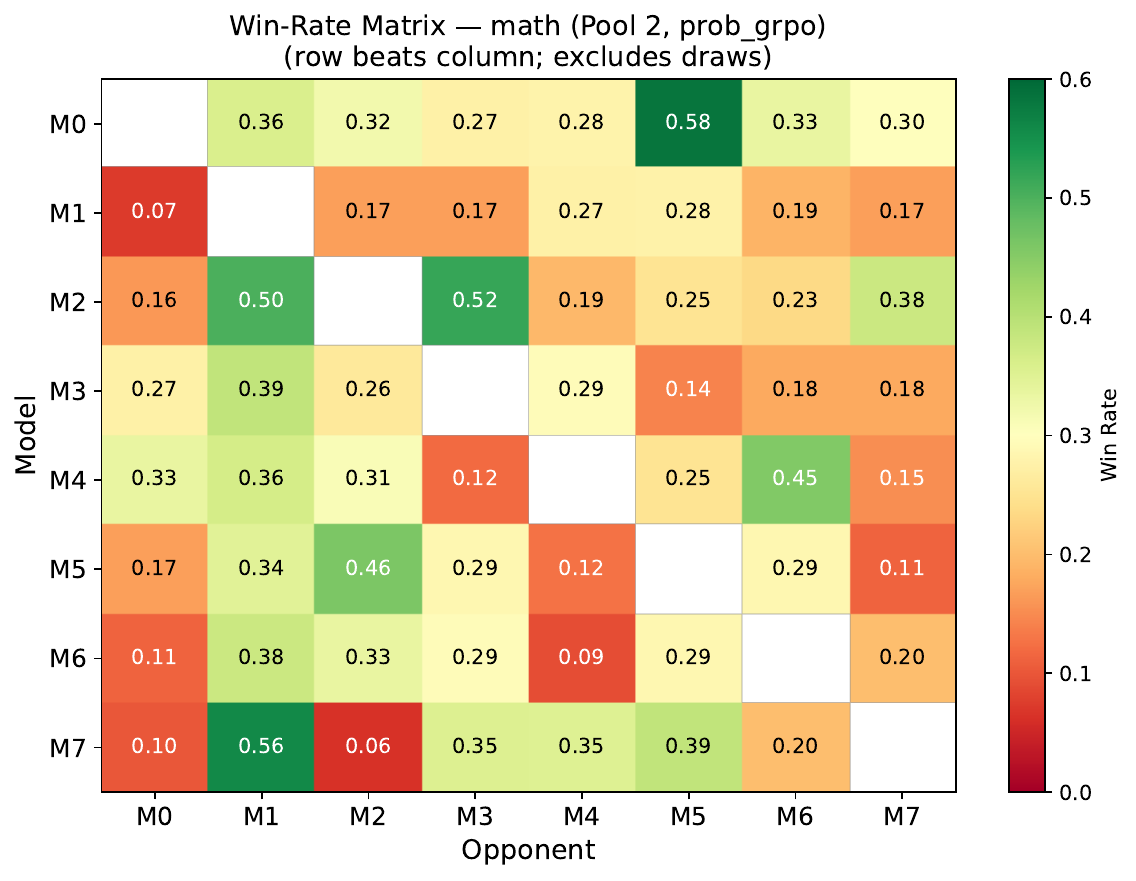}
  \caption{Pairwise win-rate matrix for the MATH dataset (Pool~2, Stackelberg GRPO). No model dominates; win rates span 0.1--0.6; $\sim$41\% of duels are draws given the objective nature of the task.}
  \label{fig:winrate_matrix}
\end{figure}

Figure~\ref{fig:winrate_matrix} shows the pairwise win-rate matrix for MATH (Pool~2, Stackelberg GRPO) referenced in \S\ref{sec:analysis}: no model dominates, win rates span 0.1--0.6, and $\sim$41\% of duels are draws given the objective nature of the task, motivating the adaptive opponent selection studied below.

Figure~\ref{fig:opponent_ablation} compares four opponent-selection strategies on the MATH task (Pool~2). All four beat the single-model baseline (0.857), confirming that competitive training helps regardless of opponent schedule. \texttt{schedule\_increasing} (0.878) outperforms the default \texttt{schedule\_decreasing} (0.865), suggesting that gradually concentrating on harder opponents is beneficial. \texttt{lowest\_diff} (0.875) also outperforms the default, while \texttt{highest\_diff} (0.861) performs worst; overexposure to very hard opponents appears noisy. Note that the three ablation variants ran for 2 iterations vs.\ 8 for the default; their final scores are not directly comparable but are reported for reference.

\subsection{Ablation: Pool Size}

\begin{figure}[t]
  \centering
  \begin{minipage}[t]{0.47\linewidth}
    \centering
    \includegraphics[width=\linewidth]{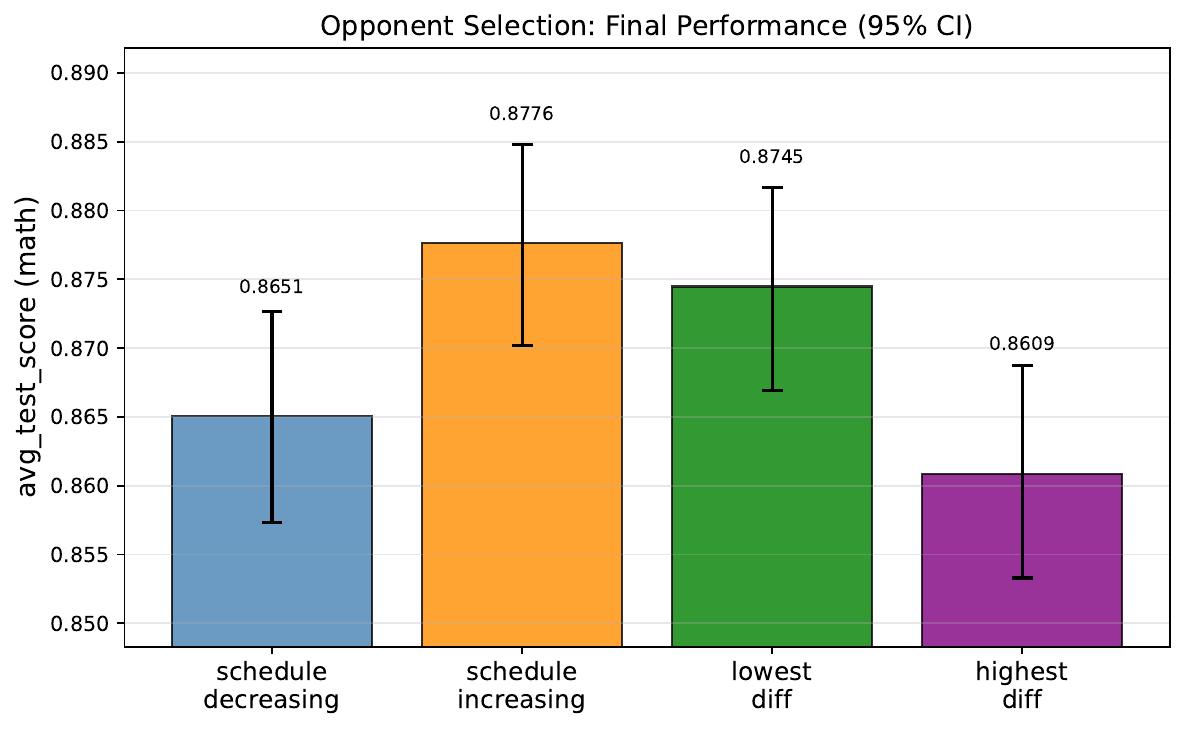}
    \caption{Opponent selection ablation on MATH (Pool~2). All strategies beat the single-model baseline (0.857); \texttt{schedule\_increasing} is best.}
    \label{fig:opponent_ablation}
  \end{minipage}
  \hfill
  \begin{minipage}[t]{0.47\linewidth}
    \centering
    \includegraphics[width=\linewidth]{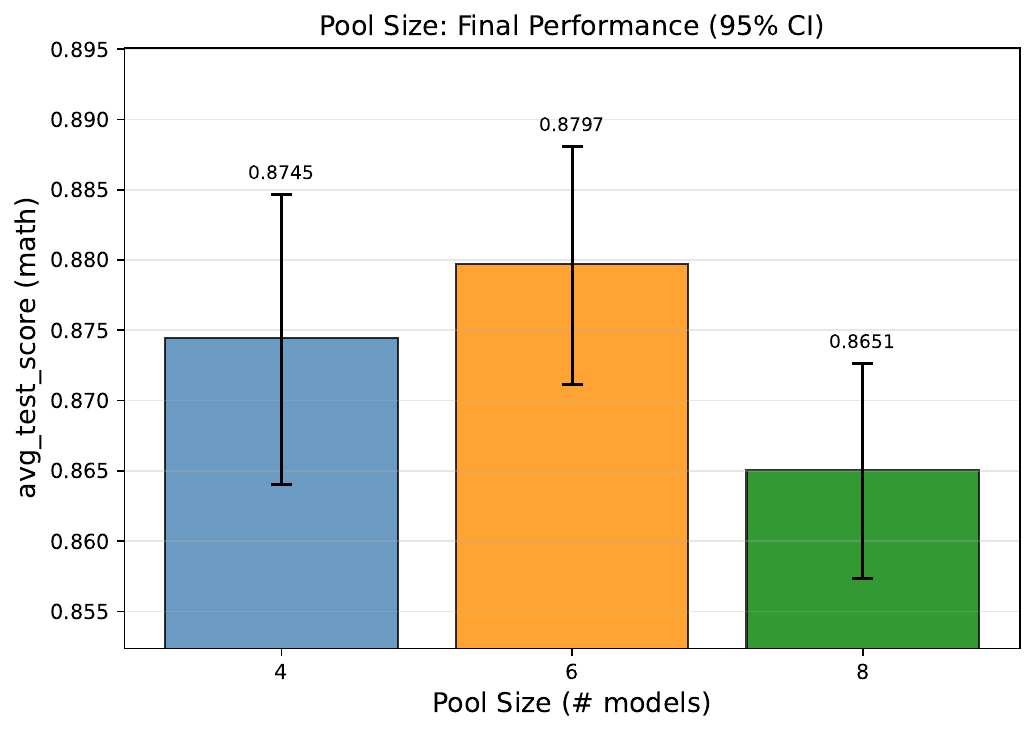}
    \caption{Pool size ablation on MATH (Pool~2). All sizes outperform the single-model baseline (0.857); Pool~6 is best.}
    \label{fig:pool_size}
  \end{minipage}
\end{figure}

Figure~\ref{fig:pool_size} compares pool sizes of 4, 6, and 8 models on the MATH task. All three beat the single-model baseline. Pool~6 (0.880) slightly outperforms Pool~4 (0.875) and Pool~8 (0.865), suggesting a moderate pool size is preferable; however, CIs overlap between Pool~4 and Pool~6. Larger pools do not monotonically help; Pool~8 underperforms both smaller configurations, consistent with the intuition that very large pools dilute the competitive signal each model receives.

\subsection{Response Diversity During Training}

\begin{figure*}[t]
  \centering
  \includegraphics[width=\linewidth]{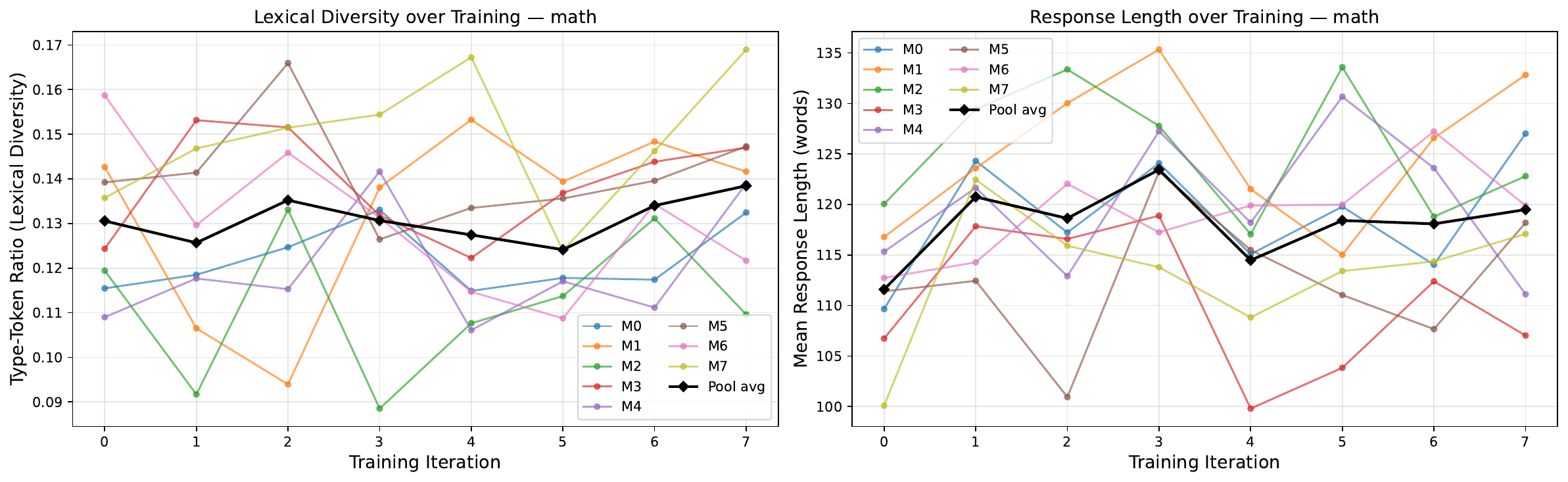}
  \caption{Response properties over training iterations (MATH, Pool~2, 8 models). Left: type-token ratio (lexical diversity). Right: mean response length. Bold line = pool average. No model collapses toward repetitive or length-degenerate outputs; pool diversity is maintained throughout.}
  \label{fig:diversity}
\end{figure*}

Figure~\ref{fig:diversity} shows lexical diversity (type-token ratio, TTR) and mean response length per model over training iterations (MATH task, Pool~2). TTR oscillates stably across all 8 iterations with no downward trend, and response length remains in the 100--135 word range throughout. Between-model variation dominates within-model trends, indicating that stylistic properties of the pre-trained models persist despite competitive training. This confirms that the pool maintains behavioral heterogeneity throughout training, a prerequisite for the leader's adaptive selection strategy to remain meaningful.

\subsection{Ablation: LLM Leader vs.\ Probabilistic Leader}
\label{app:leader_ablation}

Table~\ref{tab:leader_ablation} compares two leader strategies on Pool~1 trained with DPO and GRPO across four shared tasks.
The \emph{LLM leader} uses Gemini to select instructions based on model embeddings and current reputation scores (pilot experiment on the same Pool~1 models).
The \emph{probabilistic leader} uses EXP3, an adversarial bandit algorithm that maintains a sampling distribution over instructions updated by follower win-rates; results are taken from Table~\ref{tab:main}.

\begin{table}[h]
  \centering
  \caption{Leader strategy comparison on Pool~1 across four tasks.
    Best score per task is \textbf{bolded}.
    AlpacaEval is scored by a reward model (higher = better); all other tasks report accuracy.}
  \label{tab:leader_ablation}
  \small
  \begin{tabular}{lcccc}
    \toprule
    Method & AlpacaEval & GPQA & MBPP & TruthfulQA \\
    \midrule
    Gemini Leader + DPO   & 1.65  & 0.263 & 0.581 & 0.682 \\
    Prob.\ Leader + DPO   & \textbf{8.081} & \textbf{0.364} & 0.643 & 0.682 \\
    Prob.\ Leader + GRPO  & 7.819 & 0.353 & \textbf{0.653} & \textbf{0.684} \\
    \bottomrule
  \end{tabular}
\end{table}

The probabilistic leader substantially outperforms the Gemini leader on AlpacaEval (8.08 vs.\ 1.65) and GPQA (+10.1\% absolute), with consistent gains on MBPP (+6.2\% for DPO, +7.2\% for GRPO) and essentially matched performance on TruthfulQA. These results suggest that the expressivity of an LLM leader does not translate to better instruction selection in practice: the EXP3 bandit, which adapts directly to observed win-rates without requiring generalization from model embeddings, yields more effective curricula. The probabilistic leader also avoids the latency and cost of LLM inference at each training round. Between the two probabilistic conditions, GRPO edges out DPO on MBPP and TruthfulQA while DPO leads on GPQA, consistent with the complementary strengths in Table~\ref{tab:main}.

%%=======================================================
\section{Implementation Details}
\label{app:implementation}

\paragraph{EXP3 leader.}
We run $T{=}8$ training iterations. The leader maintains a weight vector $\mathbf{w}^t \in \mathbb{R}_{>0}^K$ over an instruction pool of size $K{=}64$.
The sampling distribution mixes exploitation and exploration with rate $\gamma{=}0.2$ (Eq.~\ref{eq:exp3-dist}).
The composite reward is $r_k = 0.3\,r_{\mathrm{diff}} + 0.7\,r_{\mathrm{pref}}$, where $r_{\mathrm{diff}}$ is the difficulty reward (Eq.~\ref{eq:diff-reward}) with floor threshold $\tau_{\mathrm{th}}{=}3.0$ (on the 1--10 judge scale), and $r_{\mathrm{pref}}$ is the preference-quality reward (Eq.~\ref{eq:pref-reward}) with Gaussian bandwidth $\sigma_r{=}0.15$ on the normalized score gap.
The ideal gap target is annealed from 0.6 to 0.15 across training.
All scores are normalized to $[0,1]$ by dividing by $s_{\max}-s_{\min}=9$.

\paragraph{Reputation system.}
The $K$ factor decays exponentially with the total number of reputation updates $n$ for each model:
\begin{equation}
K_t = \max\!\bigl(K_{\min},\; K_0 \cdot \alpha^{\lfloor n / n_s \rfloor}\bigr),
\end{equation}
with $K_0{=}10$, $K_{\min}{=}5$, decay rate $\alpha{=}0.9$, and decay step $n_s{=}10$.
This cools the magnitude of reputation swings as models accumulate more duels and their standings stabilize.
The reputation deviation $\sigma_i$ is the sample standard deviation of the last $W{=}20$ reputation deltas:
\begin{equation}
\sigma_i = \sqrt{\frac{1}{|W_i|-1}\sum_{\Delta \in W_i}(\Delta - \bar{\Delta})^2},
\end{equation}
where $W_i$ is the sliding window of the most recent $\min(n, 20)$ updates for model $M_i$ and $\bar{\Delta}$ is their mean.
$\sigma_i$ measures per-model rating volatility and enters the opponent-matching score via $z_i$ (Eq.~\ref{eq:rep-update}).

\paragraph{Opponent matching.}
We use \texttt{schedule\_decreasing} matching: the cosine-scheduled target reputation gap decreases from wide (random-like) to tight (near-peer) as training progresses.
A fraction $p_{\mathrm{rand}}{=}0.2$ of duels use uniformly random matching to maintain diversity; each combatant draws from $n_{\mathrm{opp}}{=}3$ candidate opponents.

\paragraph{DPO training.}
Each model is fine-tuned with LoRA ($r{=}64$, $\alpha{=}16$, dropout $0.1$) applied to the query, key, value, and output projections.
Optimizer: AdamW, learning rate $10^{-6}$, cosine schedule, effective batch size 16 (per-device batch 1, gradient accumulation 16), 1 epoch per iteration.
$\beta{=}0.1$, max sequence length 2048, max prompt length 1536.

\paragraph{GRPO training.}
Same LoRA configuration as DPO.
Learning rate $10^{-6}$, cosine schedule, per-device batch size 8, 1 epoch, 4 rollout generations per prompt, max completion length 256 tokens.
KL coefficient $\beta{=}0.0$; clip ratio $\epsilon{=}0.2$.
Judge rewards are normalized to $[0,1]$ via min-max scaling before being passed to the GRPO objective.

\paragraph{Generation and evaluation.}
All models generate responses with temperature $0.7$, top-$p$ $0.9$.
Peer judges generate scores greedily (temperature $10^{-5}$, max 64 tokens).
Evaluation uses task-specific formats: multiple-choice benchmarks are scored by exact match; AlpacaEval uses a reward model; open-ended scientific tasks (SMDD, AssayBench) use domain-specific metrics as described in Section~\ref{experiment}.
All baselines use identical generation and evaluation settings.

%%=======================================================
\section{Model Pool Configurations}
\label{app:pools}

Pool~1 and Pool~3 model identities are described in Section~\ref{experiment}.
Table~\ref{tab:pool2} lists the 8 models comprising Pool~2 (LLMs from Diverse Academic Research), sourced from \citet{feng2026scaling}.

\begin{table}[h]
\centering
\small
\caption{Pool~2 model configurations (LLMs from Diverse Academic Research).}
\label{tab:pool2}
\begin{tabular}{lll}
\toprule
ID & HuggingFace identifier & Specialty \\
\midrule
M0 & \texttt{chtmp223/Qwen2.5-7B-CLIPPER}                              & Reasoning/alignment \\
M1 & \texttt{chengq9/ToolRL-Qwen2.5-3B}                                & Tool use / RL \\
M2 & \texttt{AgentFlow/agentflowplanner-7b}                             & Agent planning \\
M3 & \texttt{nanami/ladder-last16L-llama3.1-8b-instruct-sft4k}         & Instruction following \\
M4 & \texttt{viswavi/qwen2.5-rlcf}                                      & RL from feedback \\
M5 & \texttt{milli19/promptmii-llama3.1-8b-instruct}                    & Prompt optimization \\
M6 & \texttt{Zhengping/conditionalprobability-regression}               & Calibration \\
M7 & \texttt{yale-nlp/MDCure-Qwen2-7B-Instruct}                        & Medical \\
\bottomrule
\end{tabular}
\end{table}

%%=======================================================
\section{Dataset Statistics}
\label{app:data}

Table~\ref{tab:data_stats} summarizes the instruction pool and test split sizes for each benchmark.
The instruction pool (Dev) is drawn from the development set of each benchmark; 80\% is used as the pool of candidate instructions available to the EXP3 leader and DPO/GRPO training, and the remaining 20\% serves as a held-out validation set for checkpoint selection.
The test set is used exclusively for final evaluation.

\begin{table}[h]
\centering
\small
\caption{Dataset sizes. Dev: full development set; 80\% forms the instruction pool for training and 20\% is held out for validation. Test: held-out evaluation set.}
\label{tab:data_stats}
\begin{tabular}{lrr}
\toprule
Benchmark & Dev & Test \\
\midrule
BixBench          & 102  & 103  \\
LabBench          & 578  & 578  \\
SMDD              & 135  & 137  \\
AssayBench        & 218  & 334  \\
GPQA-Diamond      &  99  &  99  \\
MATH              & 956  & 956  \\
HumanEval         &  50  & 114  \\
MBPP              & 487  & 487  \\
AlpacaEval        & 2000 & 1000 \\
IFEval            & 250  & 250  \\
TruthfulQA        & 200  & 617  \\
CulturalBench     & 613  & 613  \\
\bottomrule
\end{tabular}
\end{table}

%%=======================================================
\section{Baseline Configurations}
\label{app:baselines}

All baselines use the same model pools, generation hyperparameters (temperature 0.7, top-$p$ 0.9), and evaluation protocols as \ourmethod{}.
Training-based baselines (Sparta Alignment, Multiagent FT, AggLM, Trained Router) are given equal gradient updates to control for compute.

\textbf{Sparta Alignment} uses the same duel-based preference learning loop as \ourmethod{} but samples instructions uniformly from the same instruction pool, with the same opponent matching schedule and reputation system.
\textbf{Multiagent FT} collects debate-style responses from all models and fine-tunes each with supervised cross-model distillation using LoRA (lr $10^{-5}$, 3 epochs, same LoRA config as our DPO).
\textbf{AggLM} trains an RL aggregator to select among model outputs; rollout batch size 4.
\textbf{Trained Router} learns to route each query to the best model via supervised training (1 epoch, cross-entropy on dev-set oracle labels).
\textbf{Mixture of Agents (MoA)} performs one round of response synthesis (proposer $\to$ aggregator), inference only.
\textbf{Heterogeneous Swarms} runs 3 rounds of iterative refinement across the pool with population size 3.
\textbf{Multiagent Debate} uses 1 round of cross-model critique and revision.
\textbf{Majority Vote} uses plurality voting across all models at inference.

%%=======================================================
\section{Judge Prompt}
\label{app:judge_prompt}

All peer judges receive the following prompt, filled with the instruction and the response being evaluated.
Scores are parsed from JSON output and fall back to regex extraction if needed; the default score on parse failure is 5.0.

\begin{center}
\small
\begin{verbatim}
Please judge the following response based on the question
and the response to be evaluated.
Question: {instruction}
Response to be evaluated: {response}

Operation: Output ONLY a JSON object with one score in this
exact format. Score must be in the range of 1 to 10.
Your output should be like this:
{"score": score}
\end{verbatim}
\end{center}

The same prompt is used for both offline DPO judging and online GRPO reward generation.
Judge scores are aggregated as a reputation-weighted mean: $\bar{s} = \sum_{k} R_k s_k / \sum_k R_k$, where $R_k$ is the current reputation score of judge $M_k$.

%%=======================================================
\section{Computational Complexity Analysis}
\label{app:complexity}

Table~\ref{tab:complexity} summarizes the per-iteration theoretical complexity of \ourmethod{} and key baselines in terms of model forward passes $F$, training steps $S$, pool size $m$, duels per iteration $D$, instruction pool size $K$, and dataset size $N$.

\begin{table}[h]
\centering
\small
\caption{Per-iteration theoretical complexity. $m$: pool size; $D$: duels per iteration; $N$: dataset size; $K$: instruction pool size. ``Inference only'' methods have $S=0$.}
\label{tab:complexity}
\begin{tabular}{lll}
\toprule
Method & Forward passes & Training steps \\
\midrule
\ourmethod{}               & $O(D \cdot m)$ & $O(m)$ \\
Sparta Alignment           & $O(D \cdot m)$ & $O(m)$ \\
Multiagent FT              & $O(N \cdot m)$ & $O(m)$ \\
AggLM                      & $O(N \cdot m)$ & $O(1)$ \\
Trained Router             & $O(N \cdot m)$ & $O(1)$ \\
Mixture of Agents (MoA)    & $O(N \cdot m)$ & $0$ \\
Heterogeneous Swarms       & $O(N \cdot m \cdot r)$ & $0$ \\
Multiagent Debate          & $O(N \cdot m \cdot r)$ & $0$ \\
Majority Vote              & $O(N \cdot m)$ & $0$ \\
\bottomrule
\end{tabular}
\end{table}

\paragraph{Discussion.}
\ourmethod{} and Sparta Alignment share the same asymptotic complexity: each of the $D$ duels requires two generation passes (the combatants) plus $m{-}2$ judge passes, giving $O(D \cdot m)$ forward passes per iteration, followed by $m$ independent LoRA training steps.
The EXP3 leader adds only an $O(K)$ weight update per duel, which is negligible compared to model forward passes ($K{=}64$ instructions vs.\ billions of parameters per model pass).
\ourmethod{} therefore introduces \emph{zero additional model calls} relative to Sparta Alignment; all gains come from reallocating the fixed duel budget toward more informative instructions, not extra compute.

Multiagent FT, AggLM, and Trained Router each require $O(N \cdot m)$ forward passes because they process all $N$ training instructions with all $m$ models before each training update; with $N \gg D$ (e.g., $N{=}956$ vs.\ $D{=}64$ for MATH), this is substantially more expensive per iteration.
Inference-only methods (MoA, Heterogeneous Swarms, Debate, Majority Vote) pay no training cost but incur repeated forward passes at inference time; Swarms and Debate multiply by the number of refinement rounds $r$.
In our setting ($D{=}64$, $m{\le}9$), the per-iteration forward-pass count for \ourmethod{} is at most $64 \times 9 = 576$ model calls, vs.\ up to $956 \times 9 = 8{,}604$ for full-dataset sweeps.

\section{Limitations}
\label{app:limitations}

\ourmethod{} requires running all models in the pool simultaneously during the combat and judgment phases, which increases memory and compute requirements relative to single-model fine-tuning; this is a shared characteristic of all training-time multi-LLM collaboration methods.
The quality and diversity of the instruction pool $\mathcal{X}$ are important: a pool that does not span the capability frontier of the model pool limits the learning signal the EXP3 leader can identify.
As models improve across iterations, productive disagreement naturally narrows, suggesting that combining \ourmethod{} with mechanisms for dynamic instruction augmentation or pool expansion is a promising direction for future work.
Finally, the reputation system initializes all models equally and may take several iterations to reliably differentiate model strengths on highly imbalanced pools; warm-starting reputations from pre-measured benchmark performance is a natural extension.

\end{document}